\documentclass[format=acmsmall, screen=true, review=false]{acmart}
\usepackage{graphicx}
\usepackage{times}
\usepackage{xcolor}
\usepackage{soul}
\usepackage[utf8]{inputenc}
\usepackage[]{subcaption}
\usepackage{ctable}
\usepackage{booktabs}
\usepackage{multirow}
\usepackage{mathrsfs}
\usepackage{amsmath}
\usepackage{mathtools}

\newcommand{\brane}{{\textsl{Neural-Brane}}}

\usepackage{color}
\usepackage[noend]{algorithmic}
\usepackage{clrscode3e}
\usepackage[ruled]{algorithm2e}

\begin{document}
\title{Neural-Brane: Neural Bayesian Personalized Ranking for Attributed Network Embedding}
%
%\titlerunning{\brane}
\author{Vachik S. Dave and
Mohammad Al Hasan}
\affiliation{%
   \institution{Indiana University Purdue University Indianapolis, USA}
}
\email{vsdave,alhasan@iupui.edu}
\author{Baichuan Zhang}
\affiliation{%
Facebook Inc., USA}
\email{baichuan24@fb.com}
\author{Pin-Yu Chen}
\affiliation{%
IBM Research, USA}
\email{pin-yu.chen@ibm.com}

\maketitle              % typeset the header of the contribution
\begin{abstract}
Network embedding methodologies, which learn a distributed vector representation for 
each vertex in a network, have attracted considerable interest in recent years. Existing
works have demonstrated that vertex representation learned through an embedding method provides superior performance in many real-world applications, such as node classification, link prediction, and community detection. However, most of the existing methods for network embedding only utilize topological information of a vertex, ignoring
a rich set of nodal attributes (such as, user profiles
of an online social network, or textual contents of a citation network), which is abundant
in all real-life networks. A
joint network embedding that takes into account both attributional and relational information
entails a complete network information and could further enrich the learned vector representations. In this work, we present \textbf{Neural-Brane}, a novel \textbf{Neural} \textbf{B}ayesian Personalized \textbf{R}anking based \textbf{A}ttributed \textbf{N}etwork \textbf{E}mbedding. For a given network, \textbf{Neural-Brane} extracts latent feature representation of its vertices using a designed neural network model that unifies network topological information and nodal attributes; Besides, it utilizes Bayesian personalized 
ranking objective, which exploits the proximity ordering between a similar node-pair and
a dissimilar node-pair. We evaluate the quality of vertex embedding produced by \textbf{Neural-Brane} by solving the node classification and clustering tasks on four real-world datasets. Experimental results demonstrate the superiority of our proposed method over 
the state-of-the-art existing methods.

\end{abstract}
\section{Introduction}

The past few years have witnessed a surge in research on embedding the vertices of a network into a low-dimensional, dense vector space. The embedded vector representation of the vertices in such a vector space enables effortless invocation of off-the-shelf machine learning algorithms, thereby facilitating several downstream network mining tasks, including node classification~\cite{Tu-ijcai-2016}, link prediction~\cite{node2vec2016}, community detection~\cite{community-aaai-17},  job recommendation~\cite{vachik-baichuan-cikm18}, and entity disambiguation~\cite{disambiguation-cikm-17}. Most existing network embedding methods, including DeepWalk~\cite{deepwalk2014}, LINE~\cite{line2015}, Node2Vec~\cite{node2vec2016}, and SDNE~\cite{SDNE2016}, utilize the topological information of a network with the rationale that nodes with similar topological roles should be distributed closely in the learned low-dimensional vector space. While this suffices for node embedding of a bare-bone
network, it is inadequate for most of today's network datasets which include useful information beyond link connectivity. Specifically, for most of the social and communication networks, a rich set of nodal attributes is typically available, and more importantly, the similarity between a pair of nodes is dictated significantly by the similarity of their attribute values. Yet, the existing embedding models do not provide a principled approach for incorporating nodal attributes into network embedding and thus fail to achieve the performance boost
that may be obtained through modeling attribute based nodal similarity. Intuitively, joint network embedding that consider both attributional and relational information could entail complementary information and further enrich the learned vector representations.

We provide a few examples from real-life networks to highlight the importance of vertex attributes for understanding the role of the vertices and to predict their interactions. For example, users on social websites contain biographical profiles like age, gender, and textual comments, which dictate who they befriend with, and what are their
common interests. In a citation network, each scientific paper is associated with a title, an abstract, and a publication venue, which largely dictates its future citation patterns. In fact, nodal attributes are specifically important when the network topology
fails to capture the similarity between a pair of nodes.
%Take online social network as an example, if two users share some common tags, they are likely to be contextually similar even if they are far away from each other based on network topology. 
For example, in academic domain,
two researchers who write scientific papers related to ``machine learning" and ``information retrieval" are not considered to be similar by existing embedding methods (say, DeepWalk or LINE) unless they are co-authors or they share common collaborators. In such a scenario, node attributes of the researchers (e.g., research keywords) are crucial for compensating for the lack of topological similarity between the researchers. In summary,  by jointly considering the attribute homophily and the network topology, more informative node representations can be expected.

%MH: We need to talk about the limitations of other works also!
Recently, a few works have been proposed which consider
attributed network embedding~\cite{Yang.ijcai2015,Huang:2017,Zhang:2017}; however, the majority of these methods use a matrix factorization approach, which suffers from some crucial limitations.
For example, earliest among these works is Text-Associated DeepWalk (TADW)~\cite{Yang.ijcai2015}, which incorporates the text features of nodes into DeepWalk by factorizing a matrix $\textbf{M}$ constructed from the summation of a set of graph transition matrices. But, SVD based matrix factorization is both time and memory consuming, which restricts TADW to scale up to large datasets. Furthermore, obtaining an accurate matrix $\textbf{M}$ for factorization is difficult and TADW instead factorizes an approximate matrix, which reduces its representation capacity. Huang et al.~\cite{Huang:2017} proposed another matrix factorization (MF) based method, known as, Accelerated Attributed Network Embedding (AANE). It suffers from the same limitation as TADW. Another crucial limitation of the above methods is that they have a design matrix which they factorize, but such a matrix cannot  deal with nodal attributes of rich types. 
In summary, the representation power of a matrix factorization based method is found to be poorer than a neural network based method, as we will show in
the experiment section of this paper. 

We found two most recent attributed network embedding methods, GraphSAGE and Graph2Gauss, which use deep neural network methods. To generate embedding of a node, GraphSAGE~\cite{NIPS2017_6703} aggregates embedding of its multi-hope neighbors using a convolution neural network model. GraphSAGE has a high time complexity, besides such ad-hoc aggregation may introduce noise which adversely affects its performance. Recently, Bojchevski et al.~\cite{g2g2018} proposed the Graph2Gauss (G2G), where they embed each node as a Gaussian distribution. G2G uses a neural network based deep encoder to process the nodal attributes and obtains an intermediate
hidden representation, which is then used to generate the mean vector and the covariance matrix of the learned Gaussian distribution of a node. As a result, in G2G's learning, the interaction between the attribute information and the topology information of a node is poor. On the other hand, the learning pipeline of  our proposed \brane\  enables effective information exchange between the attribute and topology of a node, making it much superior than G2G while learning embedding for attributed networks. 
It is worth noting that some recent works have proposed semi-supervised attributed network embedding considering the availability of node labels~\cite{HuangWSDM17,Pan:2016:TDN:3060832.3060886}, but
our focus in this work is unsupervised attributed network embedding, for which vertex labels are not available.

\textbf{Our solution and contribution:}
%\noindent {\bf Our solution and contribution.} 
In this paper, we present \brane, a novel method for attributed network embedding.
For a vertex of the input network, \brane\ infuses its network topological information and nodal attributes by using a custom neural network model, which returns a 
single representation vector capturing both the aspects of that vertex. 
The loss function of \brane\ utilizes BPR~\cite{Rendle.uai2009} to capture attribute and topological similarities between a pair of nodes in their learned 
representation vectors.  
Specifically, the BPR objective elevates the ranking of a vertex-pair having similar attributes and topology by embedding the vertices in close proximity in the representation space, in comparison to other vertex-pairs which are not similar. 
%Finally, it utilizes the mini-batch gradient descent algorithm for parameter learning. 
We summarize the key contributions of this work as follows:
\begin{enumerate}

\item We propose \brane, a custom neural network based model for learning node embedding vectors by integrating local topology structure and nodal attributes. The source code (with datasets) of the \brane\ is available at: \url{https://git.io/fNF6X}

\item \brane\ has a novel neural network architecture which enables effective mixing of attribute and structure information for learning node representation vectors capturing both the aspects of a node. Besides, it uses Bayesian personalized ranking as its objective function, which is superior than cross-entropy based objective function used in several existing network embedding works.

\item Extensive validations on four real-world datasets demonstrate that \brane\ consistently outperforms $10$ state-of-the-art methods, which results in up to $25\%$ Macro-F1 lift for node classification and more than $10\%$ NMI gain for node clustering respectively. 

%\item \brane\ is robust to the embedding dimensionality changes and has desirable convergence behavior in its formulation.

\end{enumerate}

\section{{Related Work}}

There is a large body of works on representation learning on graphs (a.k.a. network embedding). Well known among these methods are DeepWalk~\cite{deepwalk2014} and Node2Vec~\cite{node2vec2016}, both of which
capture local topology around a node through sequences of vertices obtained by uniform or biased random walk, and then use the Skip-Gram language model for obtaining the representation of each vertex. LINE~\cite{line2015} computes the similarity of a node to other nodes as a probability distribution by computing first and second order proximities, and design a KL-divergence based objective function which minimizes the divergence between empirical distribution from data and actual distribution from the embedding vectors. GraRep~\cite{GraRep2015} is a matrix factorization based approach that leverages both local and global structural information. Furthermore, a few neural network based approaches are proposed for network embedding, such as~\cite{SDNE2016,DL-AAAI-16,HNE-KDD-15}%,jietangcikm17}
.
%Network embedding for dynamic networks are also proposed~\cite{Saha.Williams.ea:18}.
Interested readers can refer to the survey articles in ~\cite{survey1,survey2}%,survey3}
, which present a taxonomy of various network embedding methods in the existing literature. 

Most of the aforementioned works only investigate the topological structure for network embedding, which is in fact only a partial view of an attributed network. To bridge this gap, a few attributed network embedding based approaches~\cite{Yang.ijcai2015,Pan:2016:TDN:3060832.3060886,Huang:2017,Zhang:2017,NIPS-17,Tang.Qu.ea:15} %,chen2016multi}
are proposed. 
The general philosophy of such works is to integrate nodal features, such as text information and user profile,  into topology-oriented network embedding model to enhance the performance of downstream network mining tasks. For example, TADW~\cite{Yang.ijcai2015} performs low-rank matrix factorization considering graph structure and text features. %However, TADW only considers the textual features associated with each node, which fails to apply TADW to handle the node attributes with rich types in general.
Furthermore, TriDNR~\cite{Pan:2016:TDN:3060832.3060886} adopts a two-layer neural networks to jointly learn the network representations by leveraging inter-node, node-word, and label-word relationships. %However, the training procedure of three components is difficult to weight and adjust. 
%\textcolor{red}{We should use one sentence to discuss the difference of \brane\ with existing methods here}
Different from the existing methods, our proposed unsupervised embedding method (\brane) utilizes a designed neural network architecture and a novel Bayesian personalized ranking based loss function to learn better network representations.

\section{Problem Statement}

Throughout this paper, scalars are denoted by lowercase alphabets (e.g., $n$). Vectors are represented by boldface lowercase letters (e.g., $\mathbf{x}$). Bold uppercase letters (e.g., $\mathbf{X}$) denote matrices, and the $i^{th}$ row of a matrix $\mathbf{X}$ is denoted as $\mathbf{x}_{i}$. The transpose of the vector $\mathbf{x}$ is denoted by $\mathbf{x}^{T}$. The dot product of two vectors is denoted by $\langle \mathbf{a}, \mathbf{b} \rangle$. $\lVert \mathbf{X} \rVert_{F}$ is the Frobenius norm of matrix $\mathbf{X}$. Finally calligraphic uppercase letter (e.g., $\mathcal{X}$) is used to denote a set and $|\mathcal{X}|$ is used to denote the cardinality of the set $\mathcal{X}$.

\begin{figure} [t]
\centering
	\includegraphics[width=0.52\textwidth,keepaspectratio]{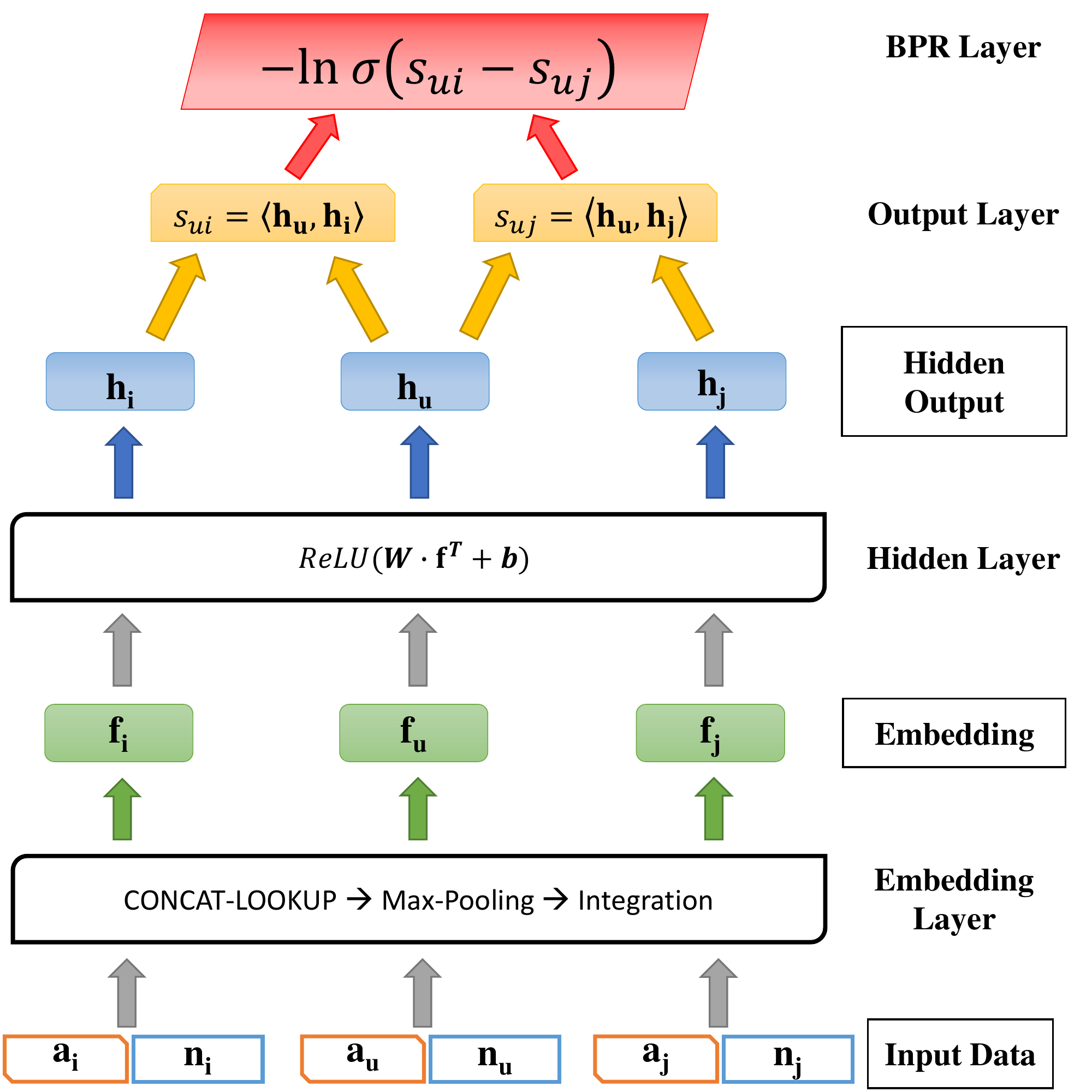}
\caption{\brane\ architecture. Given a node $u$, $\mathbf{a}_{u}$ is its binary attribute vector and $\mathbf{n}_{u}$ is its adjacency vector.
Our training uses node-triplets $(u, i, j)$, such that $(u, i) \in \mathcal{E}$ and $(u, j) \not \in \mathcal{E}$.}
\label{fig:framework}
%\vspace{-0.15in}
\end{figure}

Let $G = (\mathcal{V}, \mathcal{E}, \mathbf{A})$ be an attributed network, where $\mathcal{V}$ is a set of $n$ nodes, and $\mathcal{E}$ is a set of edges, and $\mathbf{A}$ is a $n \times m$ binary attribute matrix such that the row $\mathbf{a}_{i}$ denotes a row attribute vector associated with node $i$ in $G$.
Each edge $(i, j) \in \mathcal{E}$ is associated with a weight $w_{ij}$. The neighbors of node $i$ is represented as $\mathcal{N}(i)$. $m$ is the number of node attributes in $\mathbf{A}$. We use $\mathcal{A}(i)$ to denote the non-zero attribute set of node $i$.

The attributed network embedding problem is formally defined as follows: given an attributed network $G = (\mathcal{V}, \mathcal{E}, \mathbf{A})$, we aim to obtain the representation of its vertices as a $ n \times d$ matrix $\mathbf{F} = [\mathbf{f}_{1}^{T}, ..., \mathbf{f}_{n}^{T}]^{T} \in {\rm I \!R}^{n \times d}$, where $\mathbf{f}_{i}$ is the row vector representing the embedding of node $i$. The representation matrix $\mathbf{F}$ should preserve the node proximity from both network topological structure $\mathcal{E}$ and node attributes $\mathbf{A}$. Eventually, $\mathbf{F}$ serves as feature representation for the vertices of $G$, as such, that they can be used for various downstream network mining tasks.

\section{\brane: Attributed Network Embedding Framework}

In this section, we discuss the proposed neural Bayesian personalized ranking model for attributed network embedding. The model uses a neural network architecture with embedding layer, hidden layer, 
output layer, and BPR layer from bottom to top, as illustrated in Figure~\ref{fig:framework}. Specifically, the embedding layer learns a unified vector representation of a node from the vector representation of its nodal attributes and neighbors; the hidden layer applies nonlinear dimensionality reduction over the embedding vectors of the nodes, the output layer and the BPR layer enable model inference through back-propagation.

\label{sec:emb_layer}
\begin{figure} [!t]
\centering
	\includegraphics[width=0.52\textwidth,keepaspectratio]{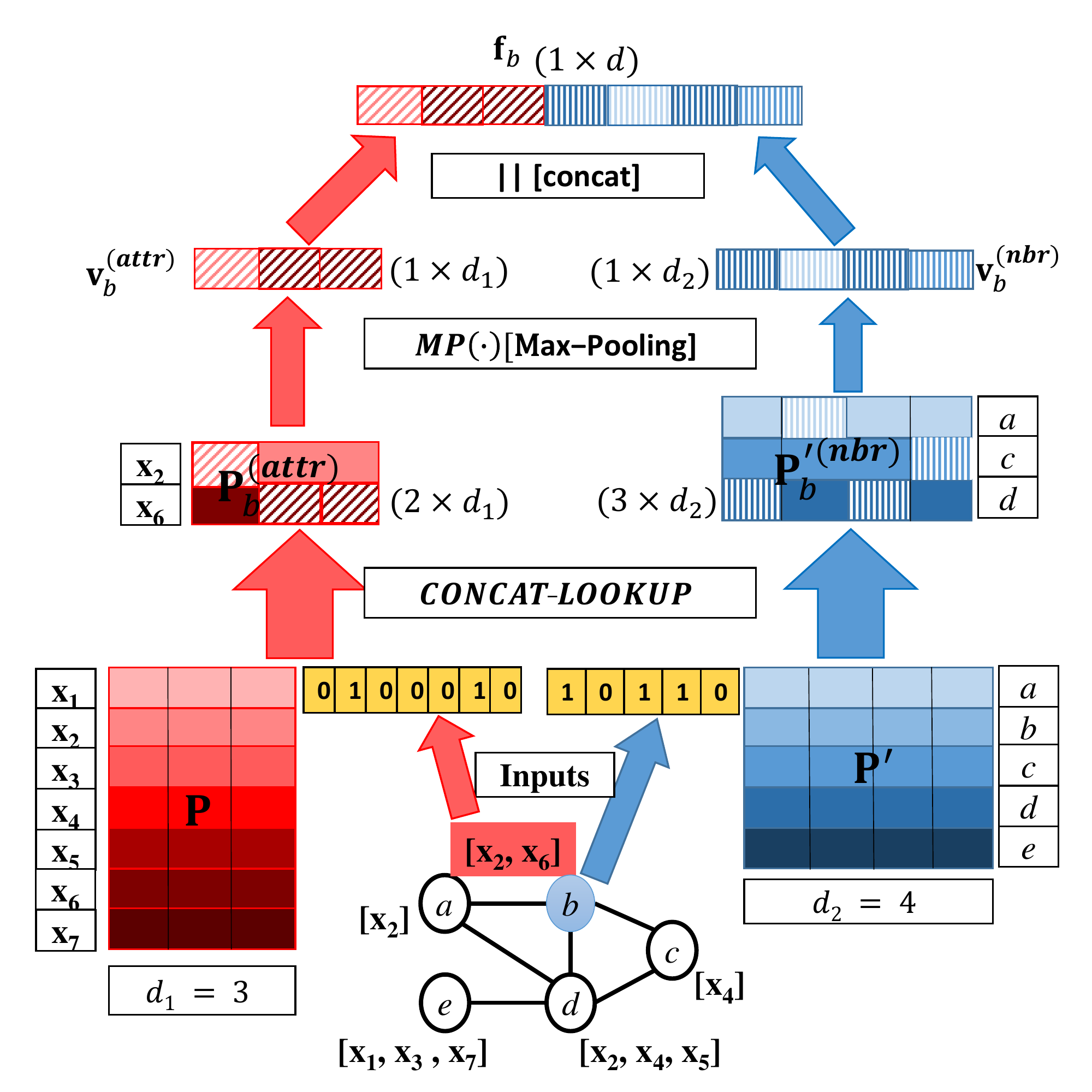}
\caption{
The figure shows the mechanism of the embedding layer for the vertex $b$ of a toy attributed graph. The graph contains 
$5$ vertices and $6$ edges, where each vertex is associated with a collection of nodal attributes.  For example, vertex $b$ is connected to vertices $\{a, c, d\}$ and associated with attributes $\{\text{x}_{2}, \text{x}_{6}\}$, respectively. The cardinality of the attribute set $\{\text{x}_1, \cdots, \text{x}_7\}$ is 7.}
\label{fig:emb_layer}
\vspace{0.1in}
\end{figure}

\subsection{Embedding Layer}
The embedding layer has two embedding matrices $\mathbf{P}$, and
$\mathbf{P}^{\prime}$; each row of $\mathbf{P}$ is a $d_1$ dimensional
vector representation of an attribute, and each row of $\mathbf{P}^{\prime}$  is a 
$d_2$ dimensional vector representation of a vertex (both $d_1$
and $d_2$ are user-defined parameter). These matrices are updated
iteratively during the learning process. For a given vertex $u$, 
embedding layer produces $u$'s latent representation vector $\mathbf{f}_u$ by learning
from embedding vectors of $u$'s attributes and neighbors, i.e., corresponding rows
of $\mathbf{P}$ and $\mathbf{P}^{\prime}$, respectively; thus the neighbors and 
attributes of 
$u$ are jointly involved in the construction of $u$'s latent representation vector ($\mathbf{f}_u$), which enables \brane\ to bring the latent representation vectors of nodes with similar attributes and neighborhood in close proximity in the latent space.

We illustrate the vector construction process using a toy attributed graph in Figure~\ref{fig:emb_layer}. Given the vertex $b$ from the toy graph, the embedding layer first takes its attribute and adjacency vectors (from $\mathbf{P}$ and $\mathbf{P}'$) as input
and then generates its corresponding attributional and nodal embedding matrices ($\mathbf{P}_b^{(attr)}$ and $\mathbf{P}_b^{\prime (nbr)}$) by using the $\small{\textit {CONCAT-LOOKUP}(\cdot)}$ function. After that, attributional and neighborhood 
embedding vectors are obtained from $\mathbf{P}_b^{(attr)}$ and $\mathbf{P}_b^{\prime (nbr)}$
by using the max-pooling operation respectively. 
Finally, the learned attributional and neighborhood embedding vectors are concatenated together to obtain the final embedding representation of the
vertex $b$. Below we provide more details of the operations in embedding
layer.

\subsubsection{Encoding attributional information.}
%\noindent{\bf Encoding attributional information.} 
Given a node $u \in \mathcal{V}$ and the attribute matrix $\mathbf{A}$, $\mathbf{a}_{u} \in {\rm I \!R}^{1 \times m}$ is $\mathbf{A}$'s
row corresponding to $u$'s binary attribute vector. We apply a row-wise concatenation based embedding lookup layer to transform $\mathbf{a}_{u}$ into a latent matrix, $\mathbf{P}_{u}^{(attr)}$, as shown below: 
\begin{equation}
\mathbf{P}_{u}^{(attr)} = \small{\textit {CONCAT-LOOKUP}}(\mathbf{P}, \mathbf{a}_{u}),
\label{eq:1}
\end{equation}
where $\mathbf{P} \in {\rm I \!R}^{m \times d_{1}}$ is the attribute embedding matrix in which each row is a $d_1$ (user defined parameter) sized vector representation of an attribute. Lookup is performed by
$\small{\textit {CONCAT-LOOKUP$(\cdot)$}}$ function which first performs a row projection on $\mathbf{P}$ by selecting
the rows corresponding to the attribute-set $\mathcal{A}(u)$ 
and then stacks the selected vectors row-wise into the matrix $\mathbf{P}_{u}^{(attr)} \in {\rm I \!R}^{|\mathcal{A}(u)| \times d_{1}}$. Then we apply a max-pooling operation on the generated $\mathbf{P}_{u}^{(attr)}$ matrix in order to transform it into a single vector. Specifically, max-pooling operation retains the most informative signal by extracting the largest value in each dimension (i.e., column) of the matrix $\mathbf{P}_{u}^{(attr)}$ to obtain $\mathbf{v}_{u}^{attr}$.

\begin{equation}
\mathbf{v}_{u}^{attr} = MP(\mathbf{P}_{u}^{(attr)}),
\label{eq:2}
\end{equation}
where $\mathbf{v}_{u}^{attr} \in {\rm I \!R}^{1 \times d_{1}}$ is the latent vector representation of node $u$ based on its attributional signals, and $MP(\cdot)$ denotes the max-pooling operation. 

\subsubsection{Encoding network topology.}
%\noindent {\bf Encoding network topology.}
Given a node $u$, we describe its neighborhood by using a binary adjacency vector, denoted as $\mathbf{n}_{u} \in {\rm I \!R}^{1 \times n}$, in which $u$'s neighbors are set to $1$, and the rest of entries are set as $0$. Similar to the operations we use for encoding the attributional information, we apply a row-wise concatenation based lookup layer to transform $\mathbf{n}_{u}$ into a latent matrix $\mathbf{P}_{u}^{\prime(nbr)}$ and then apply max-pooling operation on the obtained latent matrix. Thus,

\begin{equation}
\mathbf{P}_{u}^{\prime(nbr)} = \small{\textit {CONCAT-LOOKUP}}(\mathbf{P}^{\prime}, \mathbf{n}_{u})
\label{eq:3}
\end{equation}

\begin{equation}
\mathbf{v}_{u}^{nbr} = MP(\mathbf{P}_{u}^{\prime(nbr)}),
\label{eq:4}
\end{equation}
where $\mathbf{P}^{\prime} \in {\rm I \!R}^{n \times d_{2}}$ is the neighborhood embedding matrix for lookup (similar to matrix $\mathbf{P}$), and $\mathbf{P}_{u}^{\prime (nbr)} \in {\rm I \!R}^{|\mathcal{N}(u)| \times d_{2}}$ is the obtained latent matrix generated from the $\small{\textit {CONCAT-LOOKUP}(\cdot)}$ function. Moreover, $\mathbf{v}_{u}^{nbr} \in {\rm I \!R}^{1 \times d_{2}}$ obtained from the $MP(\cdot)$ operation is the latent vector representation of node $u$ based on its neighborhood topology. 

\subsubsection{Integration component.}
%\noindent{\bf Integration component.} 
Once we obtain the vector representation of node $u$ from both its attributional information and topological structure as developed in Equations~\ref{eq:1}, \ref{eq:2}, \ref{eq:3} and \ref{eq:4}, we further integrate both latent vectors into a unified vector representation by vector concatenation, as shown below:

\begin{equation}
\mathbf{f}_{u} = \mathbf{v}_{u}^{attr} ~||~ \mathbf{v}_{u}^{nbr} := [\mathbf{v}_{u}^{attr}~\mathbf{v}_{u}^{nbr}],
\label{eq:5}
\end{equation}
where $\mathbf{f}_{u} \in {\rm I \!R}^{1 \times d}$ ($d_{1} + d_{2} = d$), and ``$||$'' denotes the vector concatenation operation.

\subsection{Hidden Layer}

Given the obtained embedding vector $\mathbf{f_{u}}  \in {\rm I \!R}^{1 \times d}$ for node $u$ in the attributed network $G$, the hidden layer aims to transform its embedding vector into another representation $\mathbf{h}_{u}$, in which signals from
attributes and neighborhood of a vertex interact with each other. Formally, given $\mathbf{f}_{u}$, the hidden layer produces $\mathbf{h}_{u} \in {\rm I \!R}^{1 \times h}$ by the following formula:

\begin{equation}
\mathbf{h}_{u}^T = ReLU(\mathbf{W}\mathbf{f}_{u}^{T} + \mathbf{b})
\label{eq:6}
\end{equation}

Here we use rectified linear function $ReLU(x)$, defined as $\max(0, x)$,  as the activation function for achieving better convergence speed. Parameters $\mathbf{W} \in {\rm I \!R}^{h \times d}$ and $\mathbf{b} \in {\rm I \!R}^{h \times 1}$ are weights and bias for the hidden layer, respectively; $h$ is a user-defined parameter denoting the number of neurons in the hidden layer. It is worth mentioning that in the hidden layer, all the nodes share the same set of parameters $\{\mathbf{W}, \mathbf{b}\}$, which enables information sharing across different vertices (see the box denoted as ``Hidden Layer'' in Figure~\ref{fig:framework}). 

\subsection{Output and BPR Layers}

Given a node pair $u$ and $i$, we use their corresponding representations $\mathbf{h}_{u}$ and $\mathbf{h}_{i}$ from hidden layer (Equation~\ref{eq:6}) as input for the output layer. The task of this layer is to measure the similarity score between a pair of
vertices by taking the dot product of their representation vectors. Since this computation uses the vector representation of the vertices from the hidden layer, it encodes
both attribute similarity and neighborhood similarity jointly. The similarity score between vertices $u$ and $i$, defined as $s_{ui}$, is calculated as $\langle \mathbf{h}_{u}, \mathbf{h}_{i} \rangle$.

BPR layer implements the Bayesian personalized ranking objective. For the embedding task,
the ranking objective is that the neighboring nodes in the graph should have more similar vector representations in the embedding space than non-neighboring nodes. For example, the similarity score between two neighboring vertices $u$ and $i$, should be larger than the similarity score between two non-neighboring nodes $u$ and $j$. %i.e., $(u, j) \not \in \mathcal{E}$.  
As shown in Figure~\ref{fig:framework}, given the vertex triplet $(u, i, j)$, we model the probability of preserving ranking order $s_{ui} > s_{uj}$ using the sigmoid function  $\sigma(x) = \frac{1}{1 + e^{-x}}$. Mathematically,
\vspace{-0.05in}
\begin{eqnarray}
P\big(s_{ui} > s_{uj} | \mathbf{h}_{u}, \mathbf{h}_{i}, \mathbf{h}_{j}\big) &=& \sigma\big(s_{ui} - s_{uj}\big) \nonumber \\
&=& \frac{1}{1 + e^{-\big(\langle \mathbf{h_{u}}, \mathbf{h}_{i} \rangle - \langle \mathbf{h}_{u}, \mathbf{h}_{j} \rangle\big)}} \nonumber \\
\label{eq:8}
\end{eqnarray}

As we observe from Equation~\ref{eq:8}, the larger the difference between $s_{ui}$ and $s_{uj}$, the more likely the ranking order $s_{ui} > s_{uj}$ is preserved. By assuming that all the triplet based ranking orders generated from the graph $G$ to be independent, 
the probability of all the ranking orders being preserved is defined as follows:

\begin{equation}
\prod_{(u, i, j) \in \mathcal{D}} P(i >_{u} j) = \prod_{(u, i, j) \in \mathcal{D}} \sigma\big(s_{ui} - s_{uj}\big), 
\label{eq:9}
\end{equation}
where $\mathcal{D}$ represents training triplet sets generated from $G$ and $i >_{u} j$ is a shorthand notation
denoting $s_{ui} > s_{uj}$; the notation is motivated from the concept that $i$ is larger than $j$ considering the partial
order relation $>_{u}$.

The goal of our attributed network embedding is to maximize the expression in Equation~\ref{eq:9}. For the computational convenience, we minimize the sum of negative-likelihood loss function, which is shown as below:

\begin{equation}
\mathcal{L}(\Theta) = -\sum_{(u, i, j) \in \mathcal{D}} \ln \sigma\big(s_{ui} - s_{uj}\big) + \lambda \cdot ||\Theta||_{F}^{2} \\
\label{eq:10}
\end{equation}
where $\Theta = \{\mathbf{P}, \mathbf{P}^{\prime}, \mathbf{W}, \mathbf{b}\}$ are model parameters used in all different layers, and $\lambda  \cdot ||\Theta||_{F}^{2}$ is a regularization term to prevent model overfitting. 
% * <bz3@umail.iu.edu> 2018-04-08T17:13:28.446Z:
% 
% add pesdo-code 
% 
% ^.
%\vspace{-0.15in}
\subsubsection{Model inference and optimization.}
%\noindent{\bf Model inference and optimization.} 
We employ the back propagation algorithm by utilizing mini-batch gradient descent to optimize the parameters $\Theta = \{\mathbf{P}, \mathbf{P}^{\prime}, \mathbf{W}, \mathbf{b}\}$ in our model. The main process of mini-batch gradient descent is to first sample a batch of triplets from $G$. Specifically, given an arbitrary node $u$, we sample one of its neighbors $i$, i.e., $i \in \mathcal{N}(u)$, with the probability proportional to the edge weight $w_{ij}$. On the other hand, we sample its non-neighboring node $j$, i.e., $j \not \in \mathcal{N}(u)$, with the probability proportional to the node degree in the graph. Then for each mini-batch training triplets, by using the chain rule, we compute the derivative and update the corresponding parameters $\Theta$ by walking along the descending gradient direction. In particular, by back-propagating from Bayesian personalized ranking layer to hidden layer, we update the gradients w.r.t. weight matrix $\mathbf{W}$ and bias vector $\mathbf{b}$ accordingly. Then in the embedding layer, we update the gradients of the corresponding embedding vectors (i.e., rows) in $\{\mathbf{P}, \mathbf{P^{\prime}}\}$ associated with all the neighboring nodes and attributes involved in each mini-batch training triplets respectively. Mathematically, 

\begin{equation}
\Theta^{t + 1} = \Theta^{t} - \alpha \times \frac{\partial \mathcal{L}(\Theta)}{\partial \Theta}
\label{eq:11}
\end{equation}
where $\alpha$ is the learning rate. In addition, we initialize all model parameters $\Theta$ by using a Gaussian distribution with $0$ mean and $0.01$ standard deviation. 
The pseudo-code of the proposed \brane\ framework is summarized in Algorithm~\ref{alg:framework}.

\begin{algorithm}[!t]
\renewcommand{\algorithmicrequire}{\textbf{Input:}}
\renewcommand{\algorithmicensure}{\textbf{Output:}}
\caption{\brane\ Framework}
\label{alg:framework}
\begin{algorithmic}[1]
\REQUIRE $G = (\mathcal{V}, \mathcal{E}, \mathbf{A})$, embedding dimensions $d_{1}$, $d_{2}$, batch size $b$, learning rate $\alpha$, regularization coefficient $\lambda$. 
\ENSURE Attributional embedding matrix $\mathbf{P}$ and neighborhood embedding matrix $\mathbf{P}^{\prime}$. \\
%\STATE Initialize $\mathcal{L}(\Theta)^{prev}$.
\STATE Initialize all model parameters $\Theta = \{\mathbf{P}, \mathbf{P}^{\prime}, \mathbf{W}, \mathbf{b}\}$ with $0$ mean and $0.01$ standard deviation from the Gaussian distribution.
\REPEAT
\STATE Construct the mini-batch of node-triples $(u,i,j)$.
\STATE Calculate $\mathbf{f}_{u},\mathbf{f}_{i},\mathbf{f}_{j}$ using Equations~\ref{eq:1},~\ref{eq:2},~\ref{eq:3},~\ref{eq:4},~\ref{eq:5}.
\STATE Calculate $\mathbf{h}_{u},\mathbf{h}_{i},\mathbf{h}_{j}$ based on the Equation~\ref{eq:6}.
\STATE Calculate $s_{ui} = \langle \mathbf{h}_{u}, \mathbf{h}_{i} \rangle$ and
$s_{uj} = \langle \mathbf{h}_{u}, \mathbf{h}_{j} \rangle$
\STATE Calculate $\mathcal{L}(\Theta)$ using Equation~\ref{eq:10}.
\STATE Update the gradients of $\Theta = \{\mathbf{P}, \mathbf{P}^{\prime}, \mathbf{W}, \mathbf{b}\}$ using the back-propagation.
%\STATE Calculate $\Delta \mathcal{L}(\Theta) \leftarrow |\mathcal{L}(\Theta) - \mathcal{L}(\Theta)^{prev}|$
%\STATE $\mathcal{L}(\Theta)^{prev} \leftarrow \mathcal{L}(\Theta)$
\UNTIL{Convergence} %$\Delta \mathcal{L}(\Theta) < \epsilon $.}
\RETURN $\mathbf{P}, \mathbf{P}^{\prime}$.
\end{algorithmic}
\end{algorithm}

For the time complexity analysis, given the sampled training triplet set $\mathcal{D}$, the total costs of calculating and updating gradients of $\mathcal{L}$ w.r.t. corresponding embedding vectors involved in $\{\mathbf{P}, \mathbf{P^{\prime}}\}$ are $\mathcal{O}(d)$. Similarly, the total costs of computing and updating gradients of $\mathcal{L}$ w.r.t. parameters $\{\mathbf{W}, \mathbf{b}\}$ in the hidden layer are $\mathcal{O}(hd + h)$. Therefore, the total computational complexity of the proposed methodology for \brane\ is $|\mathcal{D}|*\big(\mathcal{O}(d) + \mathcal{O}(hd + h)\big)$. 
As time complexity of the \brane\ is linear to the embedding size and hidden layer dimension, it is extremely fast. For example, it takes only $10$ minutes to learn embedding for our largest dataset $Arnetminer$ (see Table~\ref{tab:dataset}).

\section{Experiments and Results}

In this section, we first introduce the datasets and baseline comparisons used in this work. Then we thoroughly evaluate our proposed \brane\ through two downstream data mining tasks (node classification and clustering) on four real-world networks, for which node attributes are available. Finally, we analyze the quantitative experimental results, investigate parameter sensitivity, convergence behavior, and the effect of pooling strategy of \brane.

\subsection{Experimental Setup}

\noindent {\bf Datasets.} We perform experiments on four real-world datasets, whose statistics are shown in Table~\ref{tab:dataset}. The largest among these networks has around 5.5K vertices, and 18K edges. Note that, publicly available networks exist, which are larger than the networks that we use in this work, but those larger networks are neither attributed nor they have class label for the vertices, so we cannot use those in our experiment. Nevertheless,
our largest dataset \textit{Arnetminer}, has more nodes, edges and attributes than datasets used by recent attribute embedding papers~\cite{Zhang:2017,Yang.ijcai2015}. More description of the datasets is given below.

\textit{CiteSeer\footnote{https://linqs.soe.ucsc.edu/data}} is a citation network, in which nodes refer to papers and links refer to citation relationship among papers. Selected keywords from the paper are used as nodal attributes. Additionally, the papers are classified into $6$ categories according to its research domain, namely Artificial Intelligence (AI), Database (DB), Information Retrieval (IR), Machine Learning (ML), Human Computer Interaction (HCI), and Multi-Agent Analysis.  

\textit{Arnetminer\footnote{https://aminer.org/topic\_paper\_author}} is a paper relation network consisting of scientific publications from $5$ distinct research areas. Specifically, we select a list of representative conferences and journals from each of them. 1) {\em Data Mining} (KDD, SDM, ICDM, WSDM, PKDD); 2) {\em Medical Informatics} (JAMIA, J. of Biomedical Info., AI in Medicine, IEEE Tran. on Medical Imaging, IEEE Tran. on Information and Technology in Biomedicine); 3) {\em Theory} (STOC, FOCS, SODA); 4) {\em Computer Vision and Visualization} (CVPR, ICCV, VAST, TVCG, IEEE Visualization and Information Visualization) 5) {\em Database} (SIGMOD, VLDB, ICDE). Authors and keywords similarity between two papers are used for building edges. Keywords from 
paper title and abstract are used as attributes.  

%\begin{table}[t!]
%\caption{Statistics of Three Real-World Datasets}
%%\vspace{-.5in}
%%\begin{table}[t!]
%\centering
%\scalebox{0.99}{
%\begin{tabular}{c c c c}
%\toprule
%Dataset  & \textit{CiteSeer} & \textit{Arnetminer} & \textit{Caltech}    \\ 
%% & & \textit{Cross-Domain} &  \\
%\midrule
%\# Nodes & $3,312$ & $5,509$  & $671$ \\
%\# Edges & $4,732$ & $17,938$ & $15,645$ \\
%\# Attributes & $3,703$ & $135,647$ & $64$   \\
%\# Classes & $6$ & $5$  & $2$ \\
%\bottomrule
%\end{tabular}}
%\label{tab:dataset}
%\vspace{-0.15in}
%\end{table}

\begin{table}[t!]
\caption{Statistics of Four Real-World Datasets}
%\vspace{-.5in}
%\begin{table}[t!]
\centering
\scalebox{1.0}{
\begin{tabular}{l c c c c}
\toprule
Dataset  & \# Nodes & \# Edges  & \# Attributes & \# Classes  \\ 
\midrule
\textit{CiteSeer} & $3,312$ & $4,732$ & $3,703$ & $6$\\
\textit{Arnetminer} & $15,753$ & $109,548$ & $135,647$ & $5$ \\
\textit{Caltech36} & $671$ & $15,645$ & $64$ & $2$ \\
\textit{Reed98} & $895$ & $17, 631$ & $64$ & $2$ \\
\bottomrule
\end{tabular}
}
\label{tab:dataset}
%\vspace{-0.15in}
\end{table}

% * <bz3@umail.iu.edu> 2018-04-10T00:09:39.843Z:
% 
% use the data link other than referece [18] in caltech 36 and reed98
% 
% ^.
{
\textit{Caltech36} and \textit{Reed98}~\cite{TRAUD2012} are two university Facebook networks. 
Specifically, each node represents a user from the corresponding university and edge represents 
user friendship.
The attributes of each node is represented by a $64$-dimensional one-hot vector based on gender, major, second major/minor, dorm/house, and year. 
%These attributes are represented as binary vectors using on-hot encoding. 
We use student/faculty status of a node as the class label.

\subsubsection{Baseline Comparison.}
%\noindent{\bf Baseline Comparison.}  
To validate the benefit of our proposed \brane, we compare it against $10$ different methods. Among all the competing methods, DeepWalk, LINE, and Node2Vec are topology-oriented network embedding approaches. NNMF, DeepWalk + NNMF, GraphSAGE, PTE-KL, TADW, AANE and G2G are state-of-the-arts for combining both network structure and nodal attributes for network representation learning. Note that PTE-KL is a semi-supervised embedding approach, and we hold the label information out for a fair comparison. 

\begin{enumerate}

\item  \textbf{DeepWalk}~\cite{deepwalk2014}: It utilize Skip-Gram based language model to analyze the truncated uniform random walks on the graph.

\item  \textbf{LINE}~\cite{line2015}: It embeds the network into a latent space by leveraging both first-order and second-order proximity of each node.

\item  \textbf{Node2Vec}~\cite{node2vec2016}: Similar to DeepWalk, Node2Vec designs a biased random walk procedure for network embedding.

\item  \textbf{Non-Negative Matrix Factorization (NNMF):} The model captures both node attributes and network structure to learn topic distributions of each node.

\item  \textbf{DW+NNMF:} It simply concatenates the vector representations learned by DeepWalk and NNMF.

\item \textbf{GraphSAGE}~\cite{NIPS2017_6703}: GraphSAGE presents an inductive representation learning framework that leverages node feature
information (e.g., text attributes) to efficiently generate node embeddings in the network.

\item  \textbf{PTE-KL}~\cite{Tang.Qu.ea:15}: Predictive Text Embedding framework aims to capture the relations of paper-paper and paper-attribute under matrix factorization framework. The objective is based on KL-divergence between empirical similarity distribution and embedding similarity distribution.

\item  \textbf{TADW}~\cite{Yang.ijcai2015}: Text-associated DeepWalk combines the text features of each node with its topology information and uses the MF version of DeepWalk. 

\item  \textbf{AANE}~\cite{Huang:2017}: Accelerated Attributed Network Embedding learns low-dimensional representation of nodes from network linkage and content information through a joint matrix factorization.

\item \textbf{G2G}~\cite{g2g2018}: Graph2Gauss learns node representation such that each node vector is a Gaussian distribution.

\end{enumerate}

\begin{table*}[!t]
\caption{Quantitative results of Macro-F1 between our proposed \brane\ and other baselines for the node classification task using logistic regression on various datasets (embedding dimension = 150). [$^*$GraphSAGE for Arnetminer is not able to complete after $2$ days.]}
\centering
\label{table:class_results}
%\resizebox{\columnwidth}{!}{%
\scalebox{0.75}{
\renewcommand{\arraystretch}{1.0}
\begin{tabular}{l c c c | c c c c c c c | c }
\toprule
%\hline
\multicolumn{10}{c}{\textit{Citeseer}} \\
\hline
Train$\%$ & DeepWalk & LINE & Node2Vec & NNMF & DW+NNMF & GraphSAGE & PTE-KL & TADW & AANE & G2G & \brane\ \\
\hline
%$10\%$ & $0.4557$ & $0.3883$ & $0.4817$ & $0.3649$ & $0.4311$ & $0.4831$ & $0.4899$ & $0.5338$ &  & $\mathbf{0.6229}_{\pm.0097}$ \\
$30\%$ & $0.4952$ & $0.4304$ & $0.5462$ & $0.4367$ & $0.5185$ & $0.4418$ & $0.5456$ & $0.5756$ & $0.5684$ & $0.5860$ & $\mathbf{0.6375}_{\pm.0075}$ \\
$50\%$ & $0.5199$ & $0.4590$ & $0.5632$ & $0.4619$ & $0.5598$ & $0.4621$ & $0.5647$ & $0.5900$ & $0.5844$ & $0.5939$ & $\mathbf{0.6450}_{\pm.0026}$ \\
$70\%$ & $0.5318$ & $0.4600$ & $0.5743$ & $0.4711$ & $0.5780$ & $0.4662$ & $0.5732$ & $0.6106$ & $0.5996$ & $0.6003$ & $\mathbf{0.6508}_{\pm.0115}$ \\
\hline
%\vspace{0.01 in}  
\multicolumn{10}{c}{\rule{0pt}{3ex}\textit{Arnetminer}} \\
\hline
Train$\%$ & DeepWalk & LINE & Node2Vec & NNMF & DW+NNMF & GraphSAGE$^*$ & PTE-KL & TADW & AANE & G2G & \brane\ \\
\hline
$30\%$ & $0.7281$ & $0.5364$ & $0.7729$ & $0.6087$ & $0.6968$ & - & $0.5341$ & $0.7969$ & $0.7902$ & $0.8062$ & $\mathbf{0.8693}_{\pm.0016}$ \\
$50\%$ & $0.7336$ & $0.5422$ & $0.7837$ & $0.6541$ & $0.7016$ & - & $0.5426$ & $0.8031$ & $0.8009$ & $0.8145$ & $\mathbf{0.8713}_{\pm.0017}$ \\
$70\%$ & $0.7389$ & $0.5485$ & $0.7877$ & $0.6748$ & $0.7044$ & - & $0.5519$ & $0.8079$ & $0.8065$ & $0.8186$ & $\mathbf{0.8759}_{\pm.0034}$ \\
\hline
%\vspace{0.01 in}  
\multicolumn{10}{c}{\rule{0pt}{3ex}\textit{Caltech36}} \\
\hline
Train$\%$ & DeepWalk & LINE & Node2Vec & NNMF & DW+NNMF & GraphSAGE & PTE-KL & TADW & AANE & G2G & \brane\ \\
\hline
%$10\%$ & $0.7322$ & $0.8079$ & $0.7606$ & $0.4491$ & $0.7535$ & $\mathbf{0.8724}$ & $0.8432$ & $0.7092$ &  & ${0.8212}_{\pm.0580}$ \\
$30\%$ & $0.7824$ & $0.8023$ & $0.7859$ & $0.5243$ & $0.8480$ & $0.7233$ & $0.8701$ & $0.8748$ & $0.8527$ & $0.8523$ & $\mathbf{0.9219}_{\pm.0121}$ \\
$50\%$ & $0.7949$ & $0.8079$ & $0.8080$ & $0.5953$ & $0.8552$ & $0.7712$ & $0.8697$ & $0.8866$ & $0.8843$ & $0.8691$ & $\mathbf{0.9285}_{\pm.0134}$ \\
$70\%$ & $0.8217$ & $0.8112$ & $0.8131$ & $0.6445$ & $0.8712$ & $0.8220$ & $0.8786$ & $0.8929$ & $0.9008$ & $0.8977$ & $\mathbf{0.9456}_{\pm.0139}$ \\
\hline
\multicolumn{10}{c}{\textit{Reed98}} \\
\hline
Train$\%$ & DeepWalk & LINE & Node2Vec & NNMF & DW+NNMF & GraphSAGE & PTE-KL & TADW & AANE & G2G & \brane\ \\
\hline
%$10\%$ & $0.6947$ & $0.6938$ & $0.7206$ & $0.4860$ & $0.7623$ & $0.6127$ & $\mathbf{0.8360}$ & &  & $0.8009_{\pm.0329}$ \\
$30\%$ & $0.7662$ & $0.7195$ & $0.7682$ & $0.6472$ & $0.8055$ & $0.6325$ & $0.8333$ & $0.8460$ & $0.8285$ & $0.7515$ & $\mathbf{0.8788}_{\pm.0105}$ \\
$50\%$ & $0.7774$ & $0.7195$ & $0.7805$ & $0.7123$ & $0.8275$ & $0.7012$ & $0.8413$ & $0.8519$ & $0.8433$ & $0.7772$ & $\mathbf{0.8916}_{\pm.0176}$ \\
$70\%$ & $0.7927$ & $0.7446$ & $0.7925$ & $0.7695$ & $0.8321$ & $0.7682$ & $0.8590$ & $0.8636$ & $0.8660$ & $0.7925$ & $\mathbf{0.9033}_{\pm.0146}$ \\
\bottomrule
\end{tabular}
}
\vspace*{-4mm}
\end{table*}

\subsubsection{Parameter Setting and Implementation Details.}
%\noindent{\bf Parameter Setting and Implementation Details.} 
There are a few user-defined hyper-parameters in our proposed embedding model. We fix the embedding dimension $d = 150$ (same for all baseline methods) with $d_{1} = d_{2} = 75$. For the number of neurons in hidden layer $h$, we set it to be $150$. For the regularization coefficient $\lambda$ in the embedding model (see Equation~\ref{eq:10}), we set it as $0.00005$. In addition to that, we fix the learning rate $\alpha = 0.5$ (see Equation~\ref{eq:11}) and batch size to be $100$ during the model learning and optimization. 
%For all baseline methods, we perform grid search on the validation set for the regularization coefficient $\lambda \in \lbrace 0.01,0.001,0.0001\rbrace$ and learning rate $\alpha \in \lbrace 0.01,0.05,0.1,0.5\rbrace$. 
For baseline methods such as GraphSAGE, PTE-KL, AANE, G2G and others, we select learning rate $\alpha$ from the set $\lbrace 0.01,0.05,0.1,0.5\rbrace$\footnote{For GraphSAGE we also check smaller values of $\alpha$ i.e. $\lbrace 10^{-4},10^{-5},10^{-6}\rbrace$ as suggested in the paper~\cite{NIPS2017_6703}.}
using grid search.
Similarly for PTE-KL, TADW and other baseline methods regularization coefficient $\lambda$ is selected from the set $\lbrace 0.01,0.001,0.0001\rbrace$.
For random walk based baselines (DeepWalk and Node2Vec), we select the best walk length from the set $\lbrace 20,40,60,80\rbrace$. For the rest of hyper-parameters, we use default parameter values as suggested by their original papers. 
%For the implementation, we write our own code in Python and 
%use Tensorflow, NumPy, and scikit-learn libraries for data cleaning, linear algebra and machine learning operations. 
%source code (with datasets) is available at https://github.com/Vachik-Dave/Neural-Brane-Neural-Bayesian-Personalized-Ranking-for-Attributed-Network-Embedding.
%We run all the experiments on a 2.1 GHz Machine with 4GB memory running Linux OS.
%For both data processing and model implementation, we implement our own code in python and use numpy, tensorflow, scikit-learn, and networkx libraries for linear algebra, optimization, machine learning, and graph operations. We run all the experiments on a 2.1 GHz Machine with 8GB memory running Linux operating system.

\subsection{Quantitative Results}
\subsubsection{Node Classification.}
%\noindent{\bf Node Classification.}
\label{sec:node_clssification}
%We start by introducing the results of node classification task. 
For fair comparison between network embedding methods, we purposely choose a linear classifier to control the impact of complicated learning approaches on the classification performance.
Specifically, we treat the node representations learned by different approaches as features, and train a logistic regression classifier for multi-class / binary classification.  In each dataset, $p\% \in \{ 30\%, 50\%, 70\%\}$ of nodes are randomly selected as training set and the rest as test set. We use the widely used metric Macro-F1~\cite{Zaki.Wagner:14} for classification assessment. Each method is executed $10$ times and the average value is reported. 
For \brane, we also report standard deviation.
For better visual comparison, we highlight the best Macro-F1 score of each training ratio ($p$) with bold font. 

Table~\ref{table:class_results} shows results for node classification, where each column is an embedding method and rows represent different train splits ($p$). 
As we observe from Table~\ref{table:class_results}, performance of the last four (PTE-KL, TADW, AANE, G2G) baseline methods  are highly competitive among each others. But, our proposed \brane\ consistently outperforms all these and other baseline methods under all training ratios. Moreover, the overall performance improvement that our \brane\ delivers over the second best method is significant. For example, in Citeseer dataset, when training ratio $p$ ranges from $30\%$ to $70\%$, \brane\ outperforms the G2G by $8.8\%$, $8.6\%$, $8.4\%$ in terms of Macro-F1, respectively. Furthermore, the improvement over G2G is statistically significant (paired t-test with p-value $\ll$ $0.01$). The relatively good performance of our proposed \brane\ across various training ratios is due to the fact that our proposed neural Bayesian personalized ranking framework is able to generate high-quality latent features by capturing crucial ordering information between nodes and incorporating nodal attributes and network topology into network embedding. Furthermore, BPR is shown to be better suited than other loss functions, such as point-wise square loss in TADW and K-L divergence based objective in LINE and PTE-KL, for placing similar nodes in the embedding space for the downstream node classification task. %To verify the effectiveness of BPR loss in our framework, we compare it with using the log-loss (cross entropy loss) in Neural-Brane and find that BPR loss indeed yields better performance.  Similarly, we also verify that in Neural-Brane using max-pooling gives better results than using sum-pooling. The details are given in the supplementary material.

\begin{figure} [!t]
\centering
\begin{subfigure}[h]{0.32\textwidth}
\centering
%	\begin{minipage}{0.85\textwidth}	
	\includegraphics[width=\textwidth,keepaspectratio]{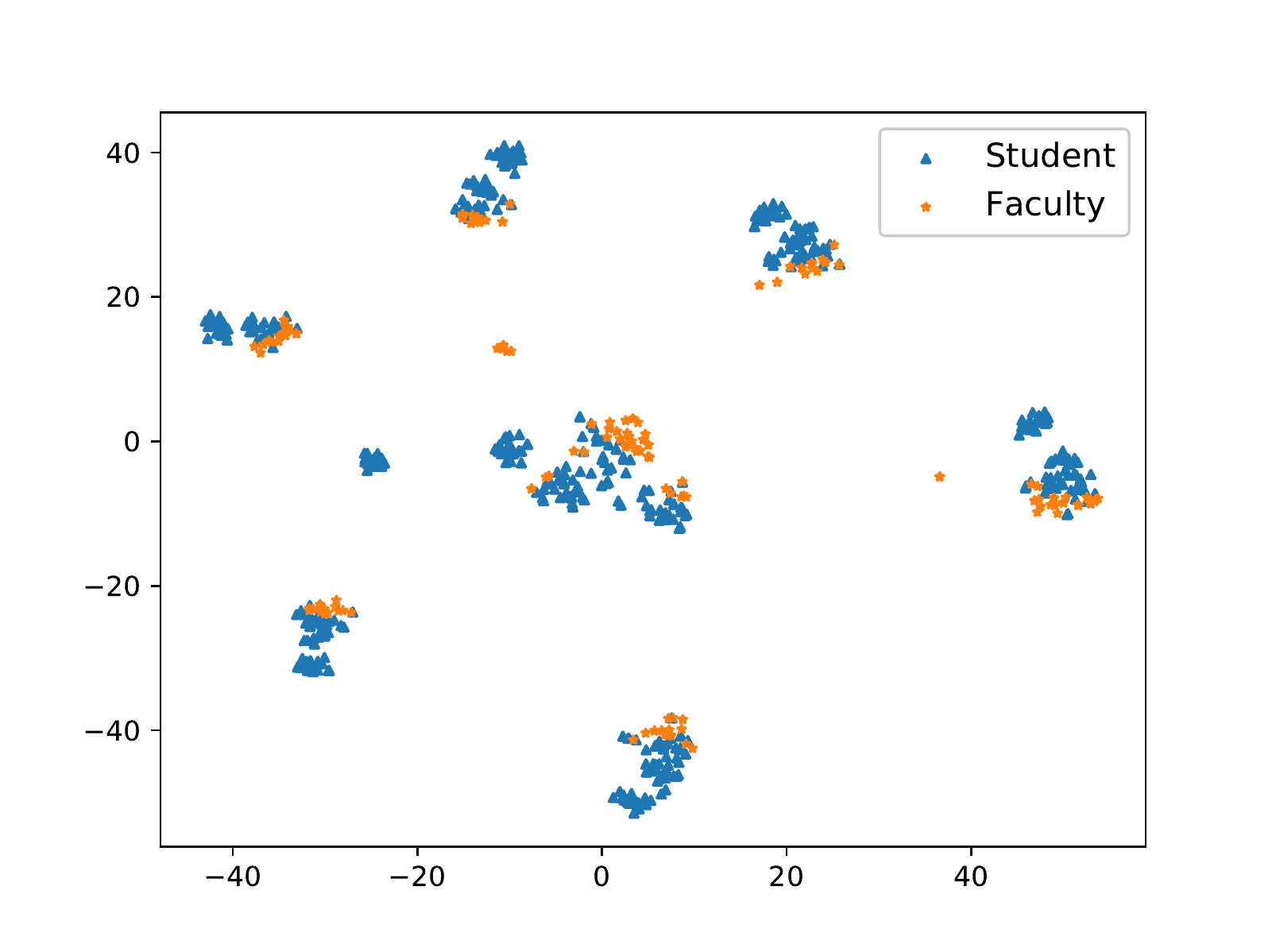}
%	\end{minipage}
	\caption{Representation of TADW for \textit{Caltech36}}
	\label{fig:Caltech_tadw}
\end{subfigure}
~
\centering
\begin{subfigure}[h]{0.32\textwidth}
\centering
%	\begin{minipage}{0.85\textwidth}	
	\includegraphics[width=\textwidth,keepaspectratio]{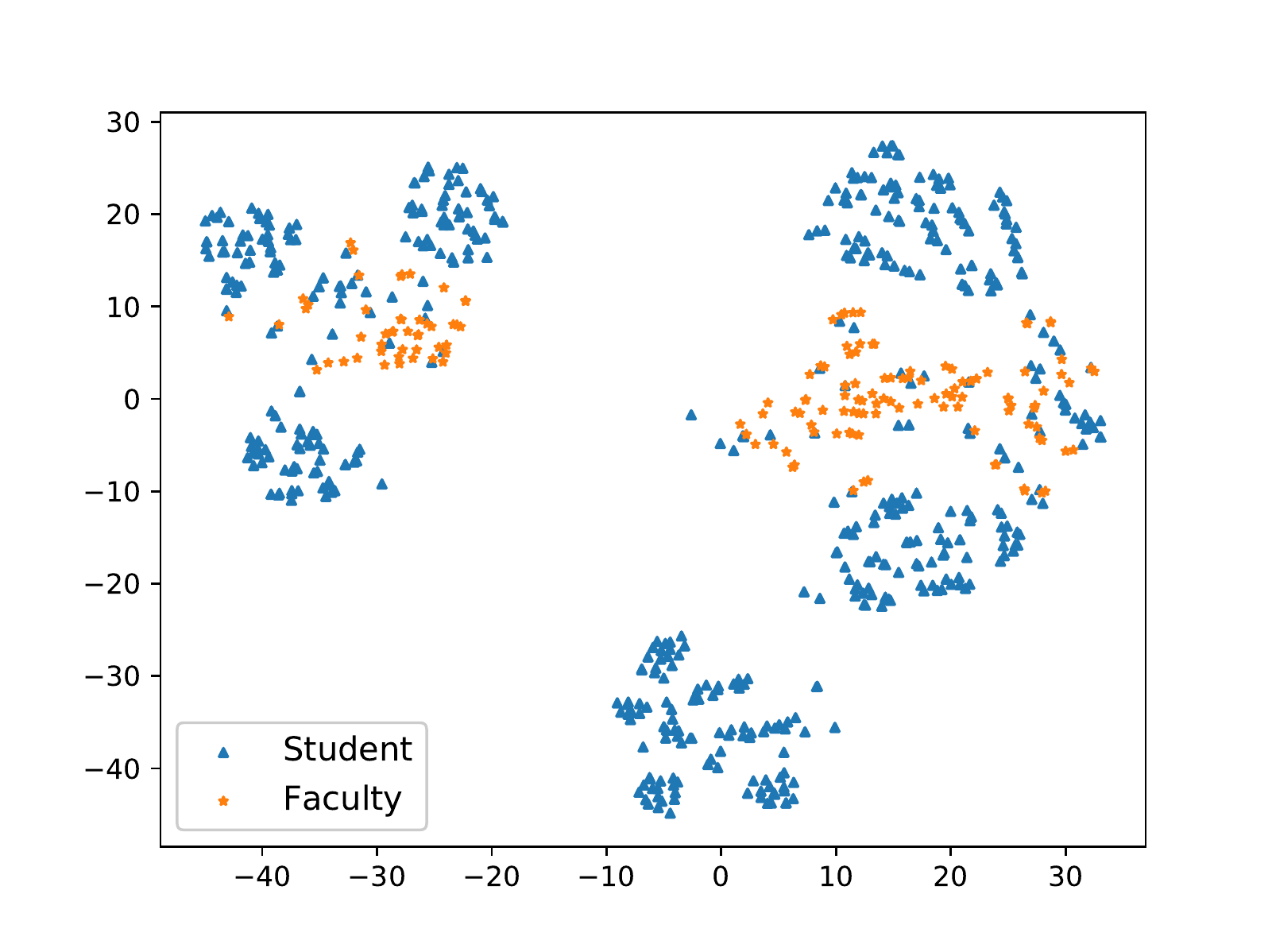}
%	\end{minipage}
	\caption{Representation of AANE for \textit{Caltech36}}
	\label{fig:Caltech_aane}
\end{subfigure}
~
\centering
\begin{subfigure}[h]{0.32\textwidth}
\centering
%	\begin{minipage}{0.85\textwidth}	
	\includegraphics[width=\textwidth,keepaspectratio]{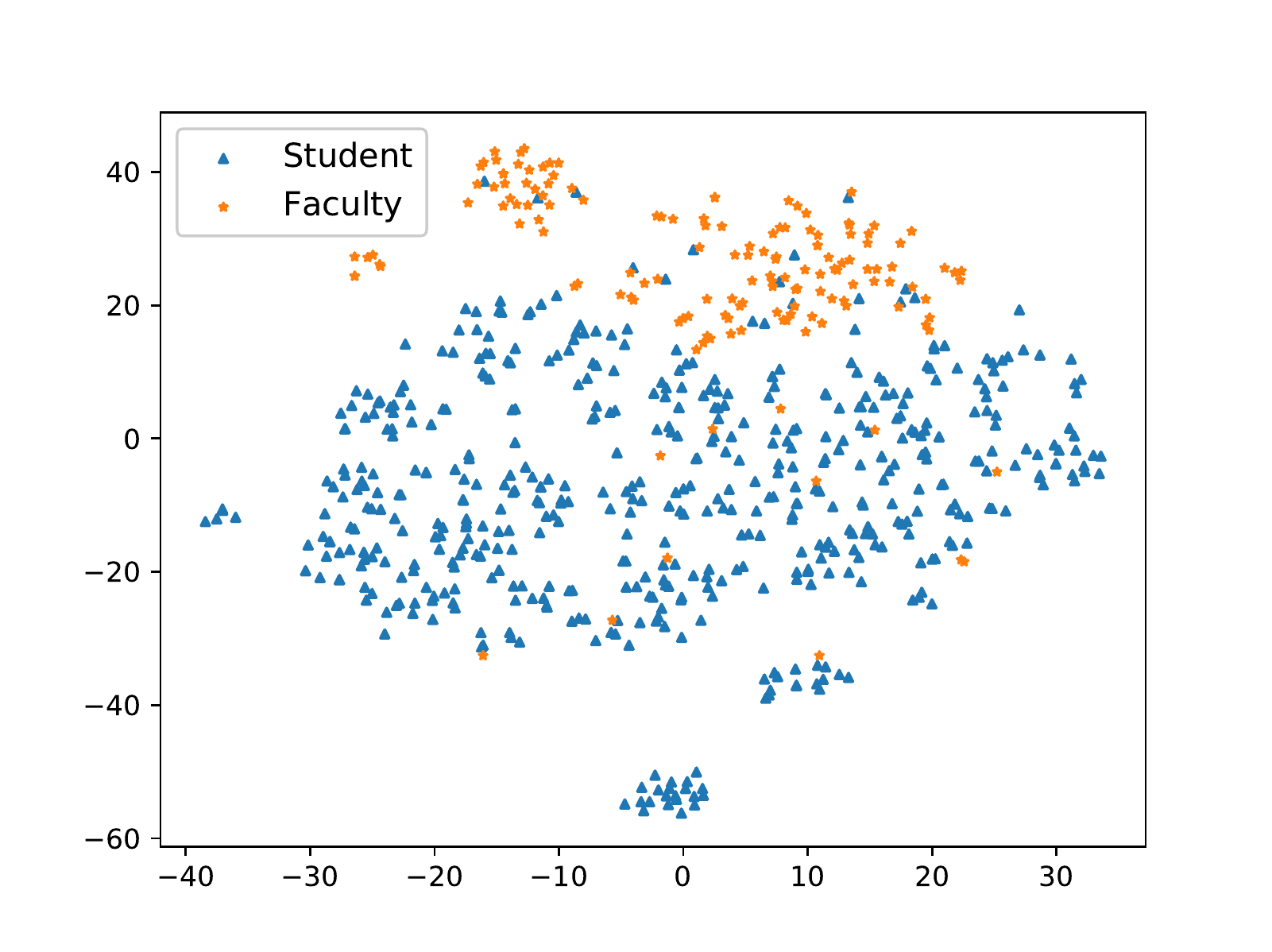}
%	\end{minipage}
	\caption{Representation of \brane\ for \textit{Caltech36}}
	\label{fig:Caltech_our}
\end{subfigure}
\centering
\begin{subfigure}[h]{0.32\textwidth}
\centering
%	\begin{minipage}{0.85\textwidth}	
	\includegraphics[width=\textwidth,keepaspectratio]{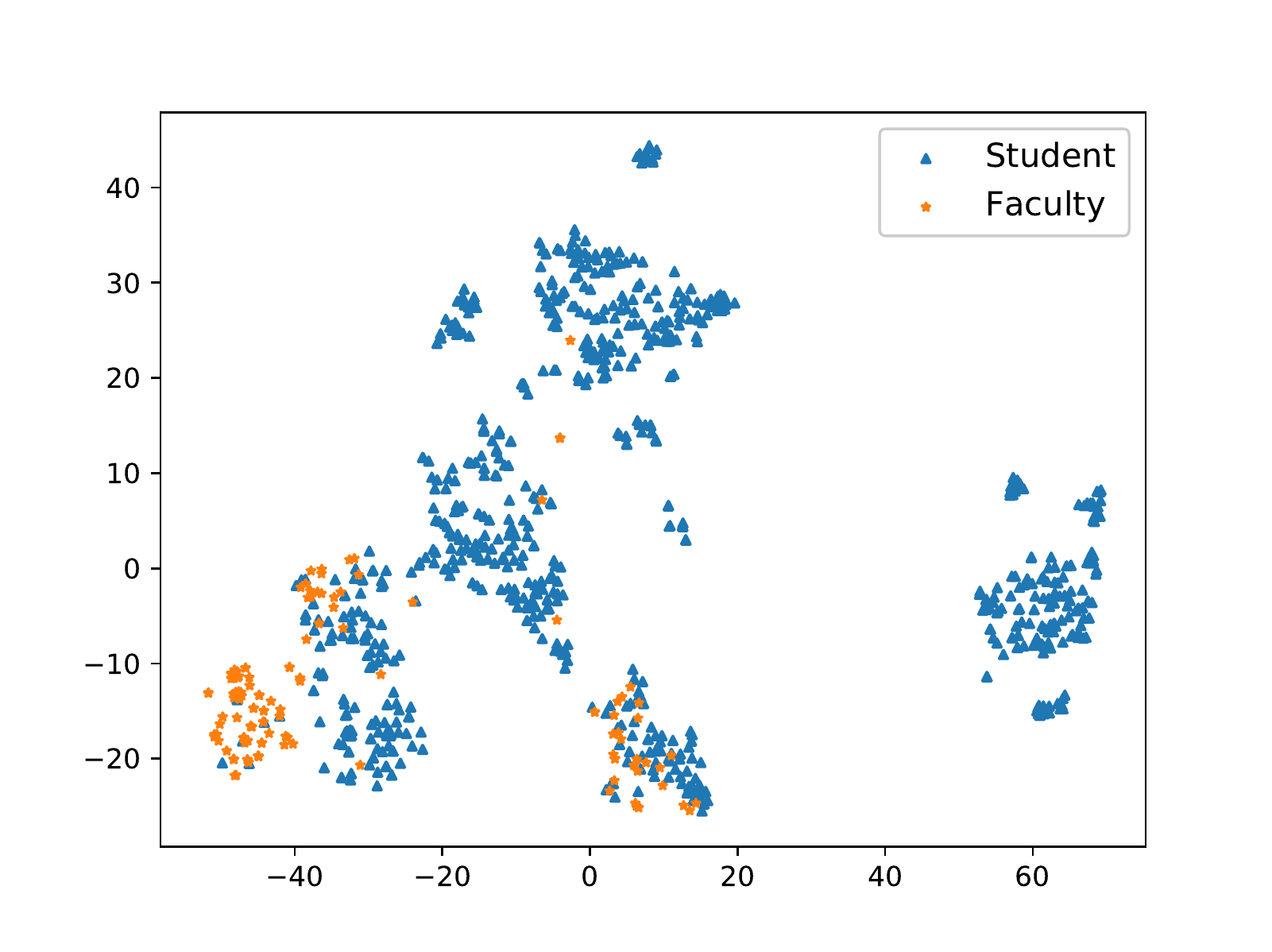}
%	\end{minipage}
	\caption{Representation of TADW for \textit{Reed98}}
	\label{fig:Reed_tadw}
\end{subfigure}
~
\centering
\begin{subfigure}[h]{0.32\textwidth}
\centering
%	\begin{minipage}{0.85\textwidth}	
	\includegraphics[width=\textwidth,keepaspectratio]{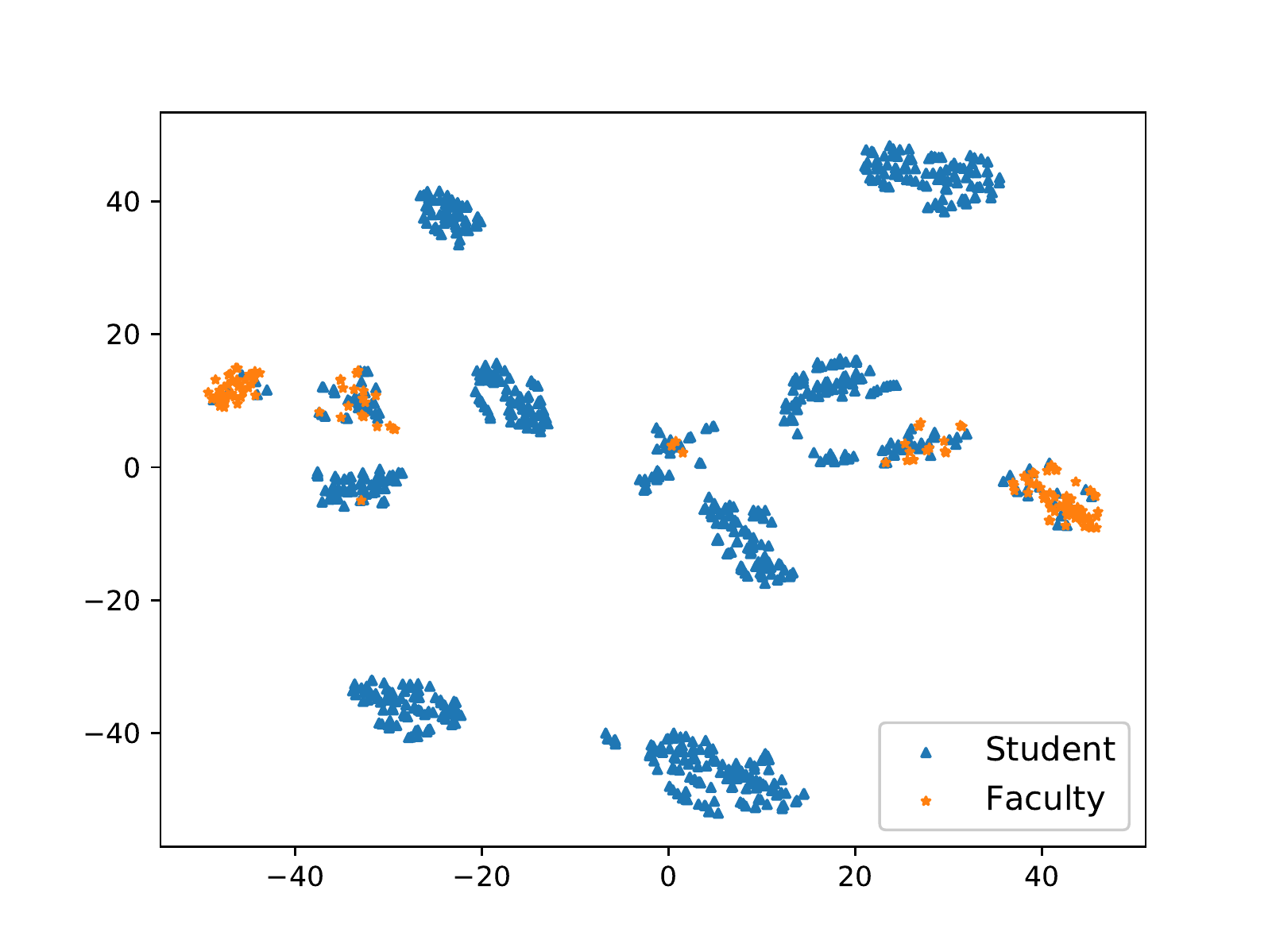}
%	\end{minipage}
	\caption{Representation of AANE for \textit{Reed98}}
	\label{fig:Reed_aane}
\end{subfigure}
%\vspace{-0.10in}
~
\centering
\begin{subfigure}[h]{0.32\textwidth}
\centering
%	\begin{minipage}{0.85\textwidth}	
	\includegraphics[width=\textwidth,keepaspectratio]{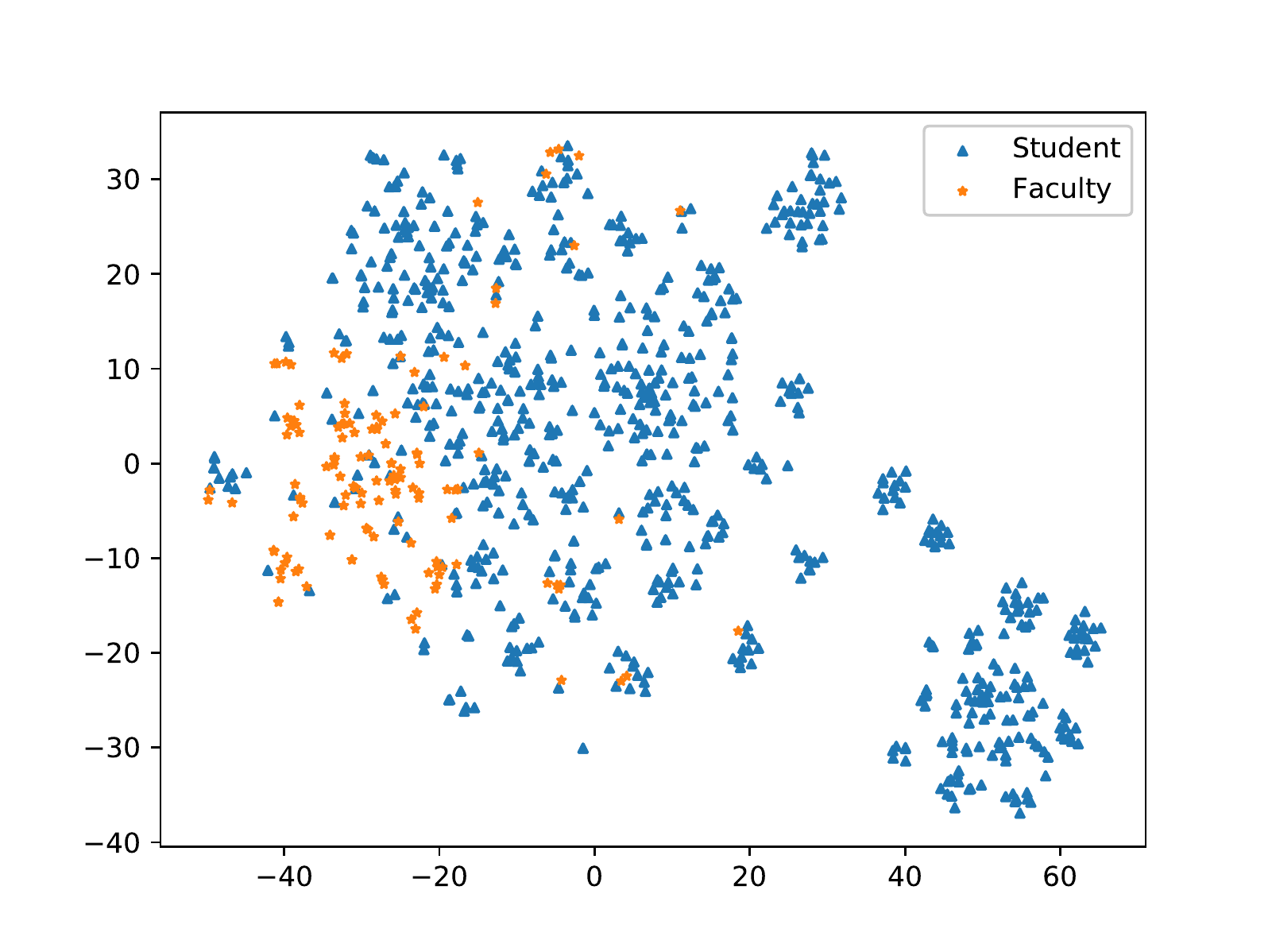}
%	\end{minipage}
	\caption{Representation of \brane\ for \textit{Reed98}}
	\label{fig:Reed_our}
\end{subfigure}
\caption{The visualization comparison among various embedding methodologies for \textit{Caltech36} and \textit{Reed98} datasets}
\label{fig:emb_plot}
%\end{minipage}
%\vspace{-0.15in}
\end{figure}

Among the competing methods, topology-oriented network embedding approaches such as LINE and DeepWalk perform fairly poor on all datasets. This is mainly because the network structure is rather sparse and only contains limited information. On the other hand, TADW is much better than DeepWalk due to the fact that textual contents contain richer signals compared to the network structure. When concatenating the embedding vectors from DeepWalk and NNMF, the classification performance is relatively improved compared to a single DeepWalk. However, the naive combination between DeepWalk and NNMF is far from optimal, compared to our proposed \brane.
Note that, GraphSAGE for Arnetminer dataset 
is not able to complete after $2$ days on contemporary server having $64$ cores with $2.3$ GHz and $132$ GB memory.

\subsubsection{Visualization and Node Clustering.}
The primary goal of graph embedding approaches is to put similar nodes closer in their corresponding latent space, hence 
a desirable embedding method should generate clusters of similar nodes in the embedding space. Visualization for large number of classes in two dimensional space is impractical. Instead, in Figure~\ref{fig:emb_plot}, we
plot $2D$ representation of learned vector representations for \textit{Caltech36} and \textit{Reed98} datasets. Note that both of these datasets contain only $2$ classes and hence provide 
interpretable visualization. Specifically, we plot embedding representations of \brane\ along with two best competing methods, namely TADW and AANE. 
%Figures~\ref{fig:Caltech_tadw},~\ref{fig:Caltech_aane},~\ref{fig:Caltech_our} illustrating $2D$
%representation of TAWD, AANE and \brane\ respectively, for \textit{Caltech36} dataset.
%Similarly, for \textit{Reed98} dataset, representation of TAWD, AANE and \brane\ are plotted 
%in the Figures~\ref{fig:Reed_tadw},~\ref{fig:Reed_aane},~\ref{fig:Reed_our}, respectively. 
These figures clearly demonstrate that \brane\ provides better discrimination of classes through clustering 
in the latent space compared to both TADW and AANE.

For the other two larger datasets (CiteSeer and Arnetminer), we use $k$-means clustering approach to the learned vector representations of nodes and utilize both Purity and Normalized Mutual Information (NMI)~\cite{Zaki.Wagner:14} to assess the quality of clustering results.  Furthermore, we match the ground-truth number of clusters as input for running $k$-means, execute the clustering process $10$ times to alleviate the sensitivity of centroid initialization, and report the average results.

%To evaluate \brane\ on the node clustering task, we apply $k$-means to the learned vector representations of nodes and utilize both Purity and Normalized Mutual Information (NMI)~\cite{Zaki.Wagner:14} to assess the quality of clustering results.  Furthermore, we match the ground-truth number of clusters as input for running $k$-means, execute the clustering process $10$ times to alleviate the sensitivity of centroid initialization, and report the average results.
% * <bz3@umail.iu.edu> 2018-08-20T00:03:56.151Z:
% 
% The statement you added "For Arnetminer dataset, clustering purity of DeepWalk is almost as good as the \brane\, however \brane\ improves ($35\%$) the performance over DeepWalk based methods in terms of NMI." are redundant
% 
% ^ <bz3@umail.iu.edu> 2018-08-20T00:46:39.977Z.
The clustering results for both \textit{CiteSeer} and \textit{Arnetminer} datasets are depicted in Figure~\ref{fig:clustering}. As we can see, our proposed \brane\ consistently achieves the best clustering results in contrast to all competing baselines. For example, in Citeseer dataset, our proposed \brane\ achieves $0.3524$ NMI. However, the best competing method PTE-KL only obtains $0.2653$ NMI, indicating more than $32.8\%$ gains. 
Similarly, for Arnetminer dataset, \brane\ obtains $34.5\%$ improvements over the best competing approach DeepWalk in terms of NMI. The possible explanation for higher performance of \brane\ could be due to the fact that our proposed Bayesian ranking formulation directly optimizes the pairwise distance between similar and dissimilar nodes, thus making their corresponding vectors cluster-aware in the embedded space. 

\begin{figure*} [!t]
\centering
\begin{subfigure}[h]{0.43\textwidth}
\centering
%	\begin{minipage}{0.85\textwidth}	
	\includegraphics[width=\textwidth,keepaspectratio]{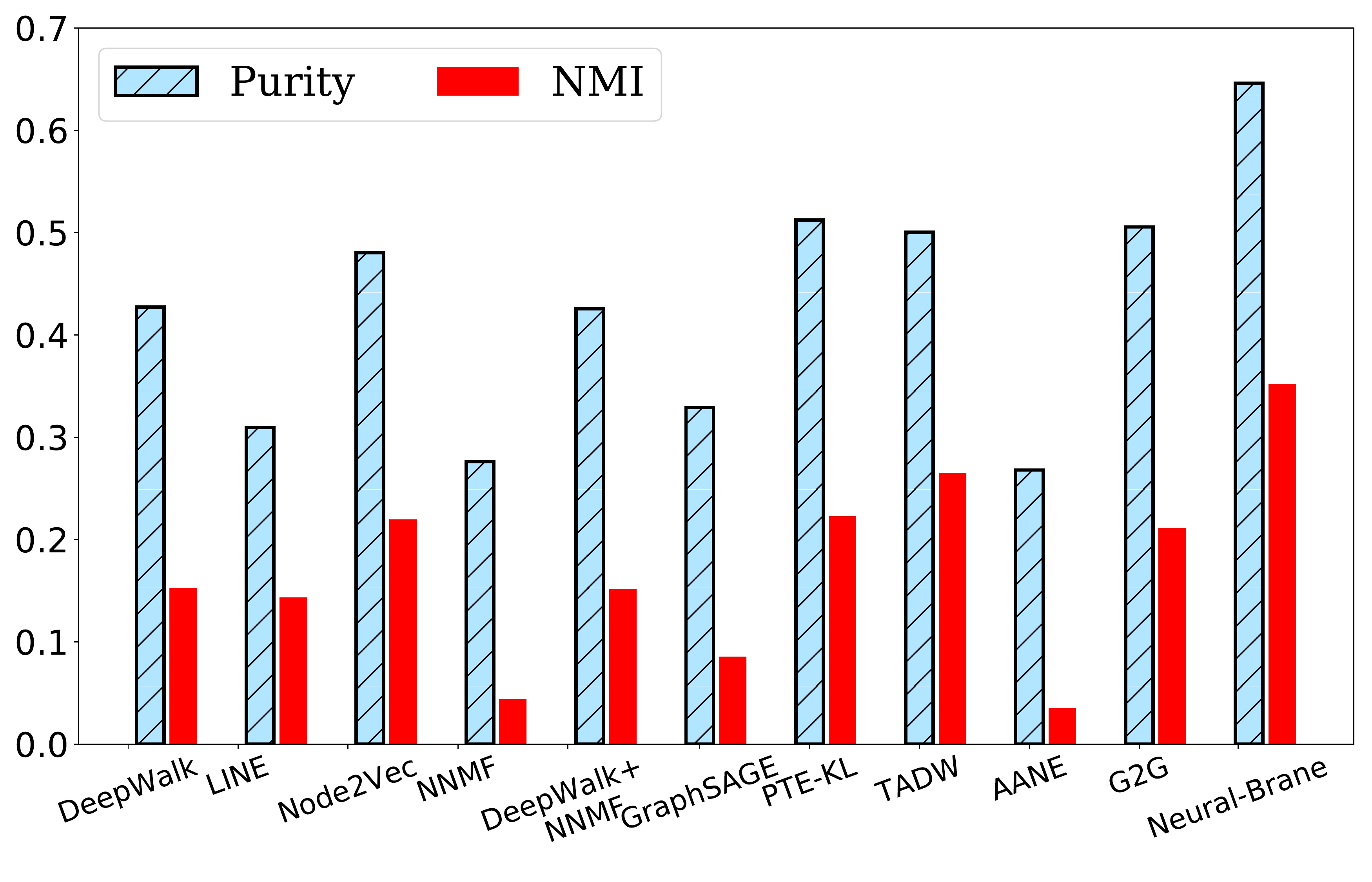}
%	\end{minipage}
	\caption{\textit{CiteSeer} Dataset}
	\label{fig:citeseer_cluster}
\end{subfigure}
~
%\vspace{0.05in}
\hspace{-0.1in}
\centering
\begin{subfigure}[h]{0.43\textwidth}
\centering
%	\begin{minipage}{0.85\textwidth}	
	\includegraphics[width=\textwidth,keepaspectratio]{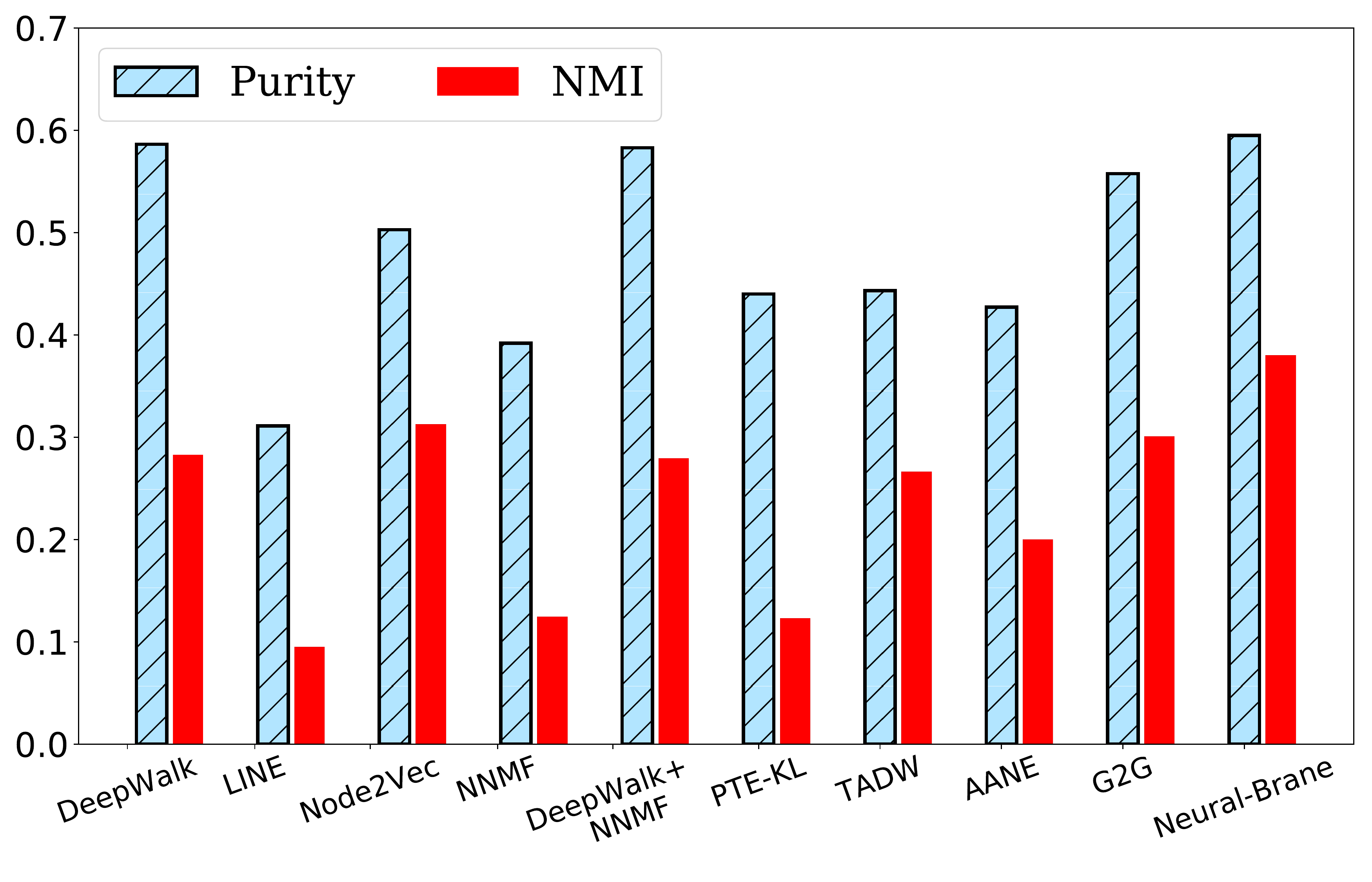}
%	\end{minipage}
	\caption{\textit{Arnetminer} Dataset}
	\label{fig:cross_cluster}
\end{subfigure}
\caption{The performance of node clustering}
\label{fig:clustering}
%\vspace{-0.15in}
\end{figure*}

\begin{figure*} [!t]
\centering
\begin{subfigure}[h]{0.40\textwidth}
\centering
%	\begin{minipage}{0.85\textwidth}	
	\includegraphics[width=\textwidth,keepaspectratio]{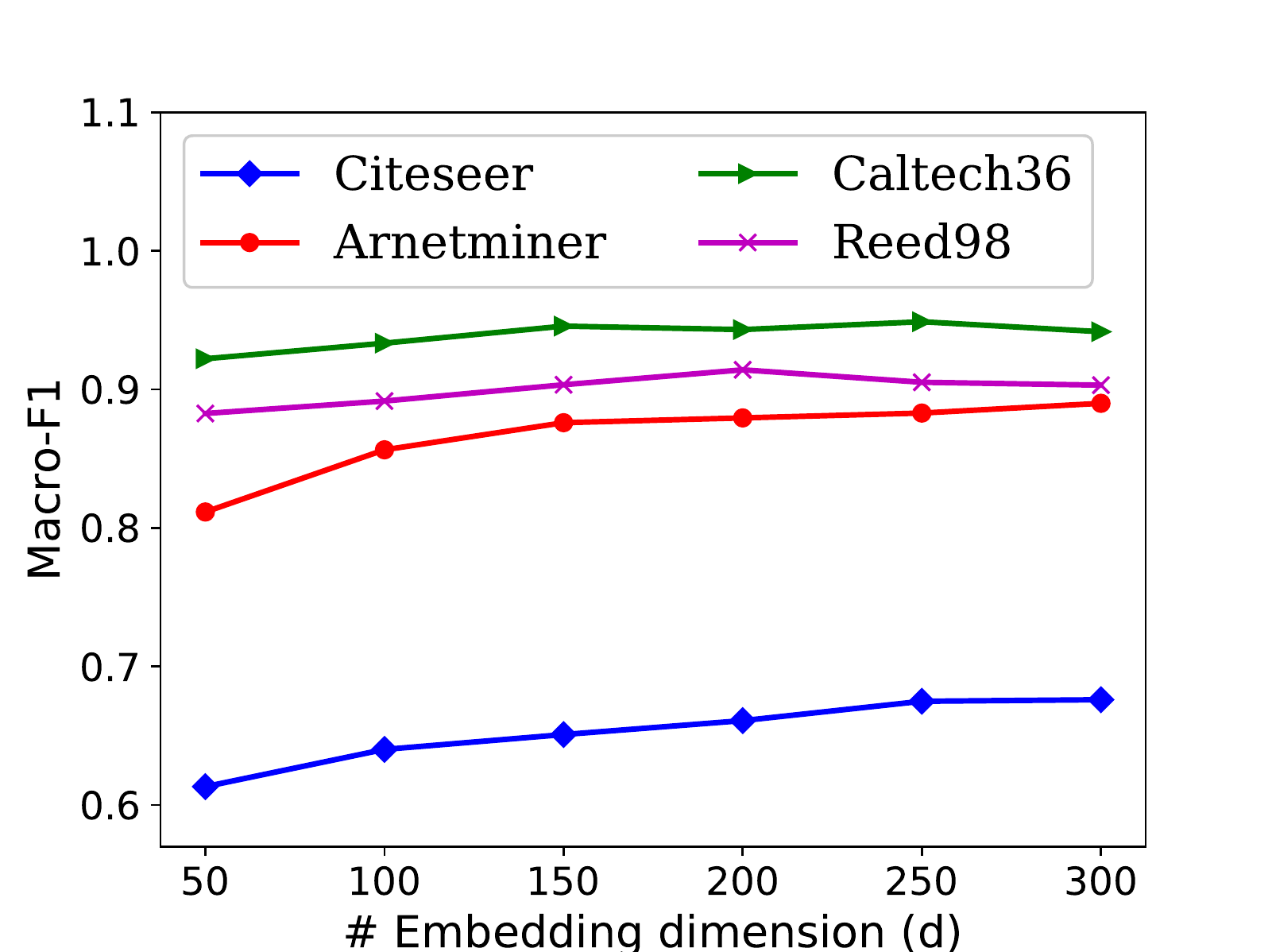}
%	\end{minipage}
	\caption{The effect of embedding dimension for node classification}
	\label{fig:para_sen}
\end{subfigure}
~
\centering
\begin{subfigure}[h]{0.40\textwidth}
\centering
%	\begin{minipage}{0.85\textwidth}	
	\includegraphics[width=\textwidth,keepaspectratio]{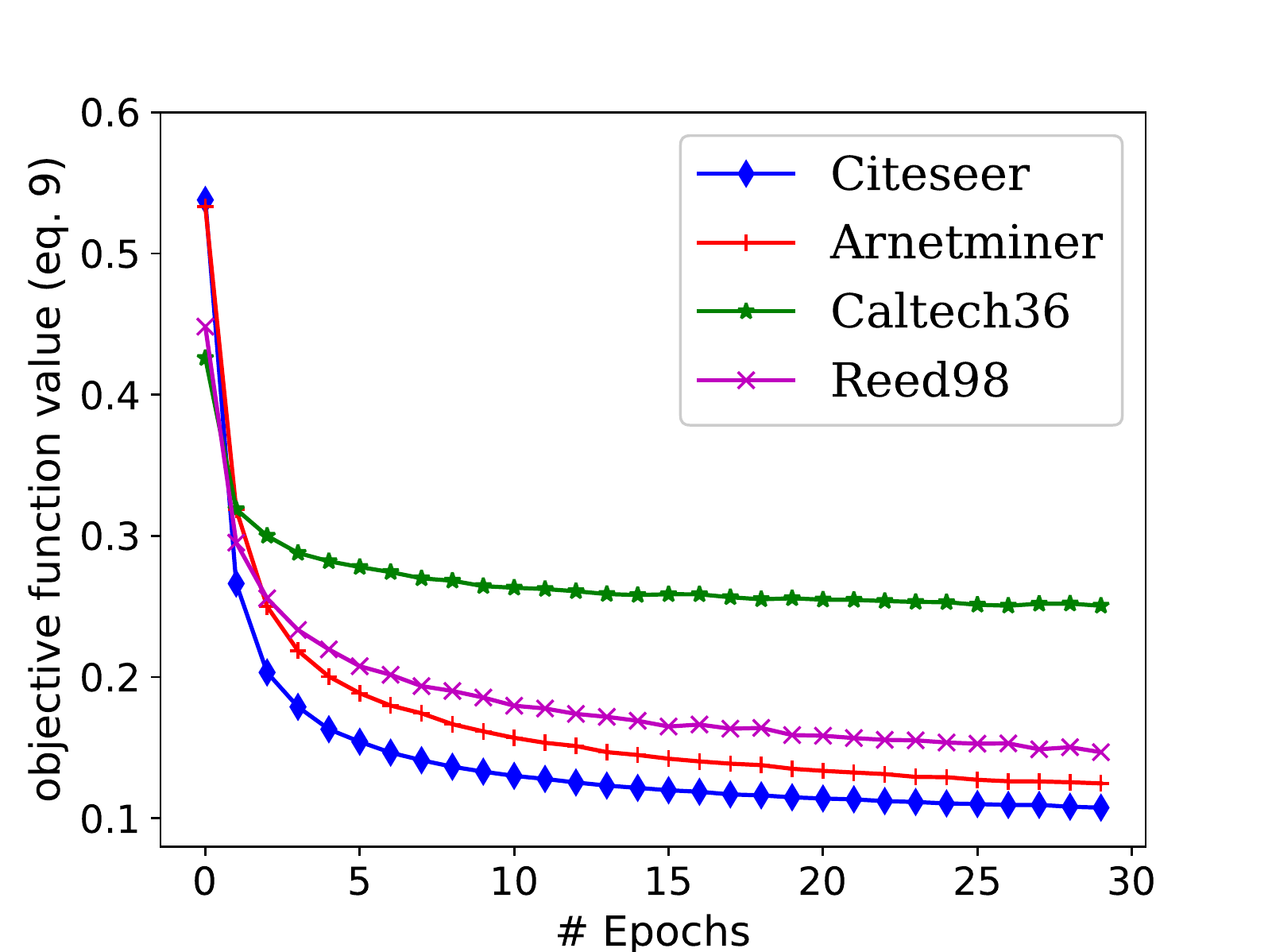}
%	\end{minipage}
	\caption{Convergence analysis of the ranking objective function shown in Equation~\ref{eq:10}}
	\label{fig:conver}
\end{subfigure}
%\vspace{-0.10in}
\caption{Analysis of the embedding dimension and convergence}
\label{fig:side_analysis}
%\vspace{-0.15in}
\end{figure*}

\subsection{Analysis of Parameter Sensitivity and Algorithm Convergence}

We conduct experiments to demonstrate how the embedding dimension affects the node classification task using our proposed \brane. Specifically, we vary the number of embedding dimension parameter $d$ as $\{50, 100, 150, 200, 250, 300\}$ and set the training ratio $p = 70\%$. We report the Macro-F1 results on all four datasets, which is shown in Figure~\ref{fig:para_sen}. As we observe, as the embedding dimension $d$ increases, the classification performance in terms of Macro-F1 first increases and then tends to stabilize. The possible explanation could be that when the embedding dimension is too small, the embedding representation capability is not sufficient. However, when the embedding dimension becomes sufficiently large, it captures all necessary information from the data, leading to the stable classification performance. Furthermore, we investigate the convergence trend of \brane. As shown in Figure~\ref{fig:conver}, \brane\ converges approximately within $10$ epochs and achieves promising convergence results in terms of the objective function value on all four datasets.

%\begin{figure}[t]
%\centering
%\includegraphics[width=55mm]{Figures/diff_d_MacroF1.eps}
%\caption{The effect of embedding dimension for the node classification performance}
%\label{fig:latent}
%\end{figure}

%\vspace{-0.1in}

\iftrue
\subsection{Effect of Pooling Strategy}
\begin{figure}[!ht]
\centering
\includegraphics[width=70mm]{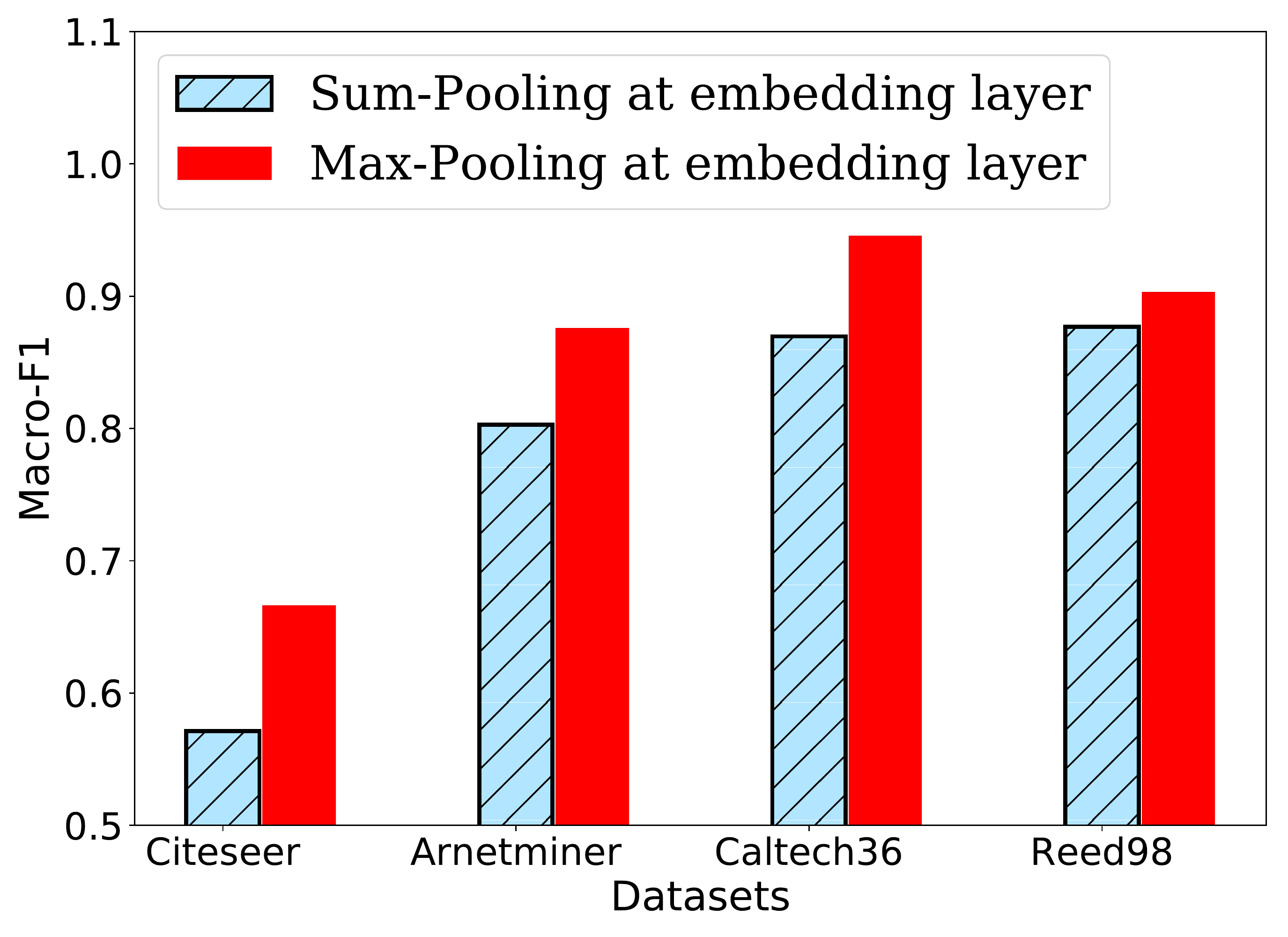}
\caption{The pooling strategy comparison for the task of node classification}
\label{fig:sumVSmax}
%\vspace{-0.2in}
\end{figure}

We finally investigate the effect of the pooling strategy in the embedding layer for the task of node classification. For the comparison, we consider taking a sum rather than the max pooling and hold the rest of neural architecture and hyper-parameter settings constant. We report the Macro-F1 results on all four datasets with training ratio $p = 70\%$, which is shown in Figure~\ref{fig:sumVSmax}.
%We discussed embedding layer mechanism in the Section~\ref{sec:emb_layer}, where we show how 
%we calculated attribute and neighborhood embedding vectors ($\mathbf{v}_u^{(attr)}$ 
%and $\mathbf{v}_u^{(nbr)}$) 
%from embedding matrices $\mathbf{P}_{u}^{(attr)}$ and $\mathbf{P}_{u}^{\prime (nbr)}$ using 
%max-pooling. 
%We can use any other pooling method for this task at 
%the embedding layer of \brane\ architecture. For this experiment replace max-pooling 
%with sum-pooling method at embedding layer and compare the performance of our model for 
%node classification task. For this comparison, we use same 
%model parameters as described in Section~\ref{sec:node_clssification} with $70\%$ training split.
As we observe, max pooling consistently performs better than alternative sum pooling
strategy for the task of node classification across all datasets.
The possible explanation is due to the fact that the max-pooling operation returns the strongest signal
for each embedding dimension, which alleviates noisy signals. On the other hand, the sum pooling operation considers accumulated
signals from each input embedding dimension, which leads to inaccurate information aggregation. 

\fi

\section{Conclusion}
We present a novel neural Bayesian personalized ranking formulation for attributed network embedding, which we call \brane. Specifically, \brane\ combines a designed neural network model and  a novel Bayesian ranking objective to learn informative vector representations that jointly incorporate network topology and nodal attributions. Experimental results on the node classification and clustering tasks over four real-world datasets demonstrate the effectiveness of the proposed \brane\ over 10 baseline methods. 

\bibliographystyle{splncs04}
\bibliography{mybib}

\begin{thebibliography}{10}
\providecommand{\url}[1]{\texttt{#1}}
\providecommand{\urlprefix}{URL }
\providecommand{\doi}[1]{https://doi.org/#1}

\bibitem{g2g2018}
Bojchevski, A., Günnemann, S.: Deep gaussian embedding of graphs: Unsupervised
  inductive learning via ranking. In: International Conference on Learning
  Representations (ICLR) (2018)

\bibitem{GraRep2015}
Cao, S., Lu, W., Xu, Q.: Grarep: Learning graph representations with global
  structural information. In: ACM International on Conference on Information
  and Knowledge Management. pp. 891--900 (2015)

\bibitem{DL-AAAI-16}
Cao, S., Lu, W., Xu, Q.: Deep neural networks for learning graph
  representations. In: AAAI. pp. 1145--1152 (2016)

\bibitem{HNE-KDD-15}
Chang, S., Han, W., Tang, J., Qi, G.J., Aggarwal, C.C., Huang, T.S.:
  Heterogeneous network embedding via deep architectures. In: International
  Conference on Knowledge Discovery and Data Mining. pp. 119--128 (2015)

\bibitem{vachik-baichuan-cikm18}
Dave, V., Zhang, B., Hasan, M.A., Jadda, K.A., Korayem, M.: A combined
  representation learning approach for better job and skill recommendation. In:
  ACM Conference on Information and Knowledge Management (2018)

\bibitem{NIPS-17}
Garc{\'{\i}}a{-}Dur{\'{a}}n, A., Niepert, M.: Learning graph representations
  with embedding propagation. In: {NIPS}. pp. 5125--5136 (2017)

\bibitem{survey2}
Goyal, P., Ferrara, E.: Graph embedding techniques, applications, and
  performance: {A} survey. CoRR  \textbf{abs/1705.02801} (2017)

\bibitem{node2vec2016}
Grover, A., Leskovec, J.: Node2vec: Scalable feature learning for networks. In:
  ACM SIGKDD International Conference on Knowledge Discovery and Data Mining.
  pp. 855--864. KDD '16 (2016)

\bibitem{NIPS2017_6703}
Hamilton, W., Ying, Z., Leskovec, J.: Inductive representation learning on
  large graphs. In: Advances in Neural Information Processing Systems 30, pp.
  1024--1034 (2017)

\bibitem{survey1}
Hamilton, W.L., Ying, R., Leskovec, J.: Representation learning on graphs:
  Methods and applications. {IEEE} Data Eng. Bull.  \textbf{40}(3),  52--74
  (2017)

\bibitem{Huang:2017}
Huang, X., Li, J., Hu, X.: Accelerated attributed network embedding. In: SIAM
  International Conference on Data Mining. pp. 633--641 (2017)

\bibitem{HuangWSDM17}
Huang, X., Li, J., Hu, X.: Label informed attributed network embedding. In: ACM
  International Conference on Web Search and Data Mining. pp. 731--739 (2017)

\bibitem{Pan:2016:TDN:3060832.3060886}
Pan, S., Wu, J., Zhu, X., Zhang, C., Wang, Y.: Tri-party deep network
  representation. In: International Joint Conference on Artificial
  Intelligence. pp. 1895--1901 (2016)

\bibitem{deepwalk2014}
Perozzi, B., Al-Rfou, R., Skiena, S.: Deepwalk: Online learning of social
  representations. In: ACM SIGKDD International Conference on Knowledge
  Discovery and Data Mining. pp. 701--710. KDD '14 (2014)

\bibitem{Rendle.uai2009}
Rendle, S., Freudenthaler, C., Gantner, Z., Schmidt-Thieme, L.: Bpr: Bayesian
  personalized ranking from implicit feedback. In: Conference on Uncertainty in
  Artificial Intelligence. pp. 452--461. UAI '09 (2009)

\bibitem{Tang.Qu.ea:15}
Tang, J., Qu, M., Mei, Q.: Pte: Predictive text embedding through large-scale
  heterogeneous text networks. In: SIGKDD. pp. 1165--1174 (2015)

\bibitem{line2015}
Tang, J., Qu, M., Wang, M., Zhang, M., Yan, J., Mei, Q.: Line: Large-scale
  information network embedding. In: International Conference on World Wide
  Web. pp. 1067--1077. WWW '15 (2015)

\bibitem{TRAUD2012}
Traud, A.L., Mucha, P.J., Porter, M.A.: Social structure of facebook networks.
  Physica A: Statistical Mechanics and its Applications  \textbf{391}(16),
  4165 -- 4180 (2012)

\bibitem{Tu-ijcai-2016}
Tu, C., Zhang, W., Liu, Z., Sun, M.: Max-margin deepwalk: Discriminative
  learning of network representation. In: {IJCAI}. pp. 3889--3895 (2016)

\bibitem{SDNE2016}
Wang, D., Cui, P., Zhu, W.: Structural deep network embedding. In: SIGKDD
  International Conference on Knowledge Discovery and Data Mining. pp.
  1225--1234. KDD '16 (2016)

\bibitem{community-aaai-17}
Wang, X., Cui, P., Wang, J., Pei, J., Zhu, W., Yang, S.: Community preserving
  network embedding. In: AAAI Conference on Artificial Intelligence (2017)

\bibitem{Yang.ijcai2015}
Yang, C., Liu, Z., Zhao, D., Sun, M., Chang, E.Y.: Network representation
  learning with rich text information. In: International Conference on
  Artificial Intelligence. pp. 2111--2117. IJCAI'15 (2015)

\bibitem{Zaki.Wagner:14}
Zaki, M.J., Jr, W.M.: Data Mining and Analysis: Fundamental Concepts and
  Algorithms. Cambridge University Press (2014)

\bibitem{disambiguation-cikm-17}
Zhang, B., Al~Hasan, M.: Name disambiguation in anonymized graphs using network
  embedding. In: ACM on Conference on Information and Knowledge Management. pp.
  1239--1248 (2017)

\bibitem{Zhang:2017}
Zhang, D., Yin, J., Zhu, X., Zhang, C.: User profile preserving social network
  embedding. In: International Joint Conference on Artificial Intelligence,
  {IJCAI-17}. pp. 3378--3384 (2017)

\end{thebibliography}

\end{document}